\documentclass[11pt,a4paper]{extarticle}
\usepackage{styles/technical_report}
\usepackage{amsmath,amsfonts,bm,mathtools,amsthm}
\usepackage{hyperref}
\usepackage{url}
\definecolor{reportReference}{HTML}{E77975}
\hypersetup{colorlinks=true,linkcolor=reportBeidaRed,citecolor=reportReference,
  urlcolor=reportReference,filecolor=reportReference,pdfborder={0 0 0},
  pdftitle={Verifier-Induced Support Reshaping in On-Policy Optimization},
  pdfauthor={Shaohang Wei, Zikun Su, Feifan Song, Wen Luo, Wei Li, Guangyue Peng, Houfeng Wang}}
\usepackage{graphicx,float,booktabs,multirow,colortbl,array,tabularx,longtable,makecell}
\usepackage{etoolbox,xparse,placeins}
\usepackage[nameinlink,noabbrev]{cleveref}
\newcommand{\papertitle}{Verifier-Induced Support Reshaping in On-Policy Optimization}
\newcommand{\ReportDate}{2026-07-31}
\newcommand{\ReportWebsiteURL}{https://sylvain-wei.github.io/VISR/}
\newcommand{\ReportCodeURL}{https://github.com/sylvain-wei/VISR}

\pdfmapline{+reportcmtt0800 SFTT0800 "T1Encoding ReEncodeFont" <cm-super-t1.enc <sftt0800.pfb}
\DeclareFontFamily{T1}{reportcmtt}{\hyphenchar\font=-1}
\DeclareFontShape{T1}{reportcmtt}{m}{n}{<->reportcmtt0800}{}
\newcommand{\ReportCorrespondenceEmail}[1]{%
  {\fontencoding{T1}\fontfamily{reportcmtt}\fontseries{m}\fontshape{n}\selectfont #1}}

\pdfmapline{+fvmr8r BeraSansMono-Roman "TeXBase1Encoding ReEncodeFont" <8r.enc <fvmr8a.pfb}
\pdfmapline{+fa5free1solid FontAwesome5Free-Solid "fa5free1 ReEncodeFont" <[fa5free1.enc <FontAwesome5Free-Solid.pfb}
\pdfmapline{+fa5brands0 FontAwesome5Brands-Regular "fa5brands0 ReEncodeFont" <[fa5brands0.enc <FontAwesome5Brands-Regular.pfb}
\DeclareFontFamily{T1}{reportlinkmono}{\hyphenchar\font=-1}
\DeclareFontShape{T1}{reportlinkmono}{m}{n}{<->fvmr8t}{}
\DeclareFontFamily{U}{reportlinkglobe}{}
\DeclareFontShape{U}{reportlinkglobe}{m}{n}{<->fa5free1solid}{}
\DeclareFontFamily{U}{reportlinkgithub}{}
\DeclareFontShape{U}{reportlinkgithub}{m}{n}{<->fa5brands0}{}
\definecolor{reportLinkBackground}{HTML}{F8F8F8}
\definecolor{reportLinkText}{HTML}{8C8E90}
\newcommand{\ReportGlobeIcon}{{\fontencoding{U}\fontfamily{reportlinkglobe}\selectfont\char128}}
\newcommand{\ReportGithubIcon}{{\fontencoding{U}\fontfamily{reportlinkgithub}\selectfont\char167}}
\newcommand{\ReportLinkButton}[3]{%
  \href{#1}{\tcbox[on line,
    colback=reportLinkBackground,colframe=reportLinkBackground,
    coltext=reportLinkText,boxrule=.4pt,arc=4pt,boxsep=2pt,
    left=4pt,right=4pt,top=2pt,bottom=2pt]{%
      \fontencoding{T1}\fontfamily{reportlinkmono}\fontseries{m}\fontshape{n}%
      \fontsize{9}{11}\selectfont #2\enspace #3}}}
\newcommand{\ReportProjectLinks}{%
  \begingroup\microtypesetup{expansion=false,protrusion=false}%
  \ReportLinkButton{\ReportWebsiteURL}{\ReportGlobeIcon}{Website}%
  \hspace{2.5pt}%
  \ReportLinkButton{\ReportCodeURL}{\ReportGithubIcon}{Code}%
  \endgroup}

\title{Verifier-Induced Support Reshaping\\in On-Policy Optimization}
\author{%
\ReportAuthor{Shaohang Wei}{1,\textdaggerdbl}, \ReportAuthor{Zikun Su}{2}, \ReportAuthor{Feifan Song}{1}, \ReportAuthor{Wen Luo}{1}, \ReportAuthor{Wei Li}{1},\\
\ReportAuthor{Guangyue Peng}{1}, \ReportAuthor{Houfeng Wang}{1,\textdagger}}
\newcommand{\ReportAffiliations}{%
\mbox{\textsuperscript{1}Peking University}\hspace{3.8em}%
\mbox{\textsuperscript{2}BUPT}}
\newcommand{\ReportAuthorNotes}{%
\mbox{\textsuperscript{\textdaggerdbl}Project Lead}\hspace{22.5pt}%
\mbox{\textsuperscript{\textdagger}Corresponding Authors}}

\newcommand{\KL}{\mathbb{D}^{\mathrm{KL}}}
\newcommand{\papertablefont}{\normalfont\rmfamily\urlstyle{same}}
\AtBeginEnvironment{table}{\papertablefont}
\AtBeginEnvironment{longtable}{\papertablefont}
\newcommand{\papergraphic}[2]{\includegraphics[width=#1]{#2}}
\DeclareMathAlphabet{\metricmathit}{OML}{cmm}{m}{it}
\DeclareRobustCommand{\metricat}[2]{%
  \ensuremath{\text{\fontencoding{OT1}\fontfamily{cmr}\mdseries\upshape\selectfont #1@}%
  \ifstrequal{#2}{k}{\metricmathit{k}}{\text{\fontencoding{OT1}\fontfamily{cmr}\mdseries\upshape\selectfont #2}}}}
\newenvironment{paperwrapfigure}[1]{%
  \begin{figure}[!htbp]\centering\begin{minipage}{0.78\textwidth}\centering
}{\end{minipage}\end{figure}}
\newenvironment{paperwraptable}[1]{%
  \begin{table}[!htbp]\centering\begin{minipage}{0.88\textwidth}\centering
}{\end{minipage}\end{table}}
\newenvironment{contributionbox}{\begin{ContributionBox}}{\end{ContributionBox}}
\NewDocumentEnvironment{keytakeaway}{O{}}{\begin{TakeawayBox}}{\end{TakeawayBox}}
\crefname{appendix}{Appendix}{Appendices}
\Crefname{appendix}{Appendix}{Appendices}
\crefname{section}{Section}{Sections}
\Crefname{section}{Section}{Sections}

\begin{document}
\maketitle
\let\ReportDefaultFootnoteRule\footnoterule
\renewcommand{\footnoterule}{\kern-3pt\hrule width12pc height0.4pt\kern2.6pt}
\begingroup
\renewcommand{\thefootnote}{\fnsymbol{footnote}}
\footnotetext[2]{%
Correspondence to \ReportCorrespondenceEmail{wanghf@pku.edu.cn}.
}
\endgroup
\addtocontents{toc}{\protect\setcounter{tocdepth}{-1}}

\begin{abstract}
    We show that on-policy reinforcement learning with verifiable rewards (RLVR) can improve the current objective while making successful behaviors for later objectives too rare to sample and reinforce.
    We call this verifier-induced support reshaping and define effective rewardable support as successful trajectories reachable within a fixed rollout budget.
    Across two model families, we study this effect through repeated verifier-scored sampling and bidirectional training on mathematical reasoning and constrained instruction following, including sequential training with the opposite verifier.
    Math-RLVR raises average instruction-following success but reduces the number of prompts with any successful response under repeated sampling.
    On IFEval with Qwen3-8B-Base, \(\metricat{pass}{1}\) rises by 6.5\% while \(\metricat{best}{32}\) falls by 9.8\%, and the same divergence appears across both models and IF benchmarks.
    Conversely, IF-RLVR shifts math responses from step-by-step openings toward direct answers, lowers \(\metricat{best}{k}\) across sampling budgets, and reduces reward variation for later Math-RLVR.
    Token-distribution analyses and controlled opening interventions show that these changes concentrate in the first few response tokens.
    RLVR mainly reranks openings already available in the base policy, and the selected opening causally affects math searchability.
    The tested reference-policy constraints, routing priors, and on-policy distillation preserve cross-task support only partially; MathIF and ReasonIF show that marginal gains translate only partly into responses that are both correct and constraint-following.
    Therefore, endpoint improvements do not guarantee future trainability or joint capability under on-policy optimization.
\end{abstract}

\begingroup
\setlength{\intextsep}{2pt}
\setlength{\abovecaptionskip}{4pt}
\begin{figure}[H]
  \centering
  \papergraphic{\linewidth}{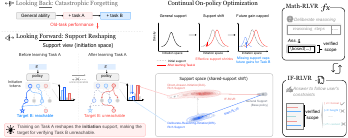}
\caption{
\textbf{Verifiers reshape what on-policy training can learn next.}
\textbf{Left:} Catastrophic forgetting looks back at retained performance; support reshaping looks ahead at future trainability.
\textbf{Middle:} Training on task A shifts task B's initiation support (top) and prompt-end hidden states (bottom).
\textbf{Right:} We test both orders with Math-RLVR and IF-RLVR.
}

  \label{fig:intro_support}
\end{figure}
\endgroup

\clearpage
\let\footnoterule\ReportDefaultFootnoteRule
\section{Introduction}

Successive post-training stages turn foundation model training into a continual-learning problem.
Alignment and reinforcement learning repeatedly adapt a shared policy to objectives such as mathematical reasoning, code generation, and instruction following \citep{ouyang2022training,shao2024deepseekmath,deepseekai2025deepseekr1,yang2024qwen25math,pyatkin2025generalizing,li2026sweext}.
Each stage should preserve both existing capabilities and the ability to learn later objectives \citep{kirkpatrick2017overcoming,parisi2019continual}.
Recent studies find that on-policy reinforcement learning often retains non-target performance better than supervised fine-tuning at comparable target-task performance \citep{shenfeld2025rlsrazor,chen2025retaining,lai2025rftforgetting}.
However, this evidence is mainly retrospective, asking what performance remains after adaptation.

Continual RLVR therefore has a forward-looking requirement: successful behaviors for later objectives must remain reachable under the current policy.
RL with verifiable rewards (RLVR) updates the policy from trajectories sampled from the current policy and scored by a verifier \citep{schulman2017proximal,shao2024deepseekmath}.
A successful behavior may remain possible yet become too rare to appear within a finite rollout budget, leaving a later objective with little positive training signal.
We call this change \emph{support reshaping} and define \emph{effective rewardable support} as reward-positive trajectories that remain likely enough to be sampled under a specified budget.
This view differs from catastrophic forgetting: forgetting measures whether performance on a learned task survives later training, whereas support reshaping measures whether successful trajectories for a future task remain discoverable before that task is optimized.
As illustrated in \Cref{fig:intro_support}, our question is not only what a verifier improves now, but also what it leaves learnable next.

We study this requirement through a controlled bidirectional comparison of mathematical reasoning and constrained instruction following.
Both are central post-training objectives with programmatically checkable rewards, but they use different success criteria.
The math verifier checks final-answer correctness while leaving the solution route open, whereas the IF verifier evaluates compliance with explicit response constraints \citep{shao2024deepseekmath,pyatkin2025generalizing}.
Prior work has also observed a tension between reasoning and instruction following at evaluation time, making this pair a useful setting for studying cross-task effects \citep{fu2025scaling,li2025thinkingfails}.
Starting from the same base policies, we train one branch with Math-RLVR and measure IF support, and train the other with IF-RLVR and measure math support.
We then continue each branch with the opposite verifier to test whether the first-stage support shift constrains later learning.
This bidirectional design reveals whether the cross-task effect changes with training direction.

Our results show that optimizing one verifier changes which successful behaviors the other verifier can still sample and reinforce.
Math-RLVR raises IF \(\metricat{pass}{1}\) but lowers \(\metricat{best}{32}\), as more prompts move toward consistent success or complete failure.
IF-RLVR lowers math \(\metricat{best}{k}\) and constrains subsequent Math-RLVR, while math responses shift from step-by-step openings toward direct answers.
Distributional analyses and controlled interventions further show that the main change occurs in the first few response tokens, where the opening choice causally affects math searchability.
In the tested settings, sequential training, reference-policy constraints, routing priors, and on-policy distillation (OPD) preserve only part of the affected support or trade preservation against target-task gains.
Results on MathIF~\citep{fu2025scaling} and ReasonIF~\citep{kwon2025reasonif} further show that gains in math or instruction following translate only partly into responses that satisfy both.

\begin{contributionbox}
\textbf{We make the following contributions:}
\begin{itemize}
    \item \textbf{Support reshaping in RLVR.} We frame continual RLVR in terms of future trainability and measure effective rewardable support through repeated verifier-scored sampling.
    Across two model families, our bidirectional Math/IF experiments show that Math-RLVR polarizes IF support, while IF-RLVR reduces math searchability and constrains subsequent Math-RLVR.
    \item \textbf{A causal role for response openings.} We localize the cross-task shift to the first few response tokens and find that RLVR mainly reranks opening options already available in the base policy.
    Controlled route interventions show that the opening choice causally affects math searchability, favoring a change in route selection over broad erasure of reasoning ability.
    \item \textbf{Limits of support preservation.} We evaluate sequential training, reference-policy constraints, routing priors, and OPD, and find that they only partially preserve future support or trade it against target-task gains.
    MathIF and ReasonIF further show that improvements in math or instruction following do not translate proportionally into responses that satisfy both.
\end{itemize}
\end{contributionbox}

\section{Related Work}

\noindent\textbf{Verifiable Rewards and Reasoning Post-training.}
Reinforcement learning with verifiable rewards (RLVR) has become an important post-training approach for improving large language models on mathematics, coding, and general reasoning tasks \citep{song2025mitigating}.
Recent reasoning systems show that executable or otherwise checkable rewards can substantially improve performance on the task being optimized \citep{shao2024deepseekmath,deepseekai2025deepseekr1,kimiteam2025kimi,yang2024qwen25math}.
Earlier work on learned verifiers, self-consistency, mathematical reasoning benchmarks, and process supervision helped establish how reasoning outputs can be evaluated and improved \citep{cobbe2021training,wang2022selfconsistency,hendrycks2021math,lightman2023lets,wei2025time}.
Overall, these works focus mainly on performance on the task currently being optimized.
We instead ask which trajectories remain available after a verifier improves its target task.
Later on-policy training can observe, reward, and reinforce only trajectories sampled from the current policy.

\noindent\textbf{Instruction Following and Reasoning.}
Instruction-following (IF) evaluation has shifted from broad preference judgments to reproducible checks of explicit constraints.
Verifiable instruction-following benchmarks turn length, format, keyword, and structural requirements into automatically checkable constraints, and recent work shows that these rewards can directly improve instruction following \citep{zhou2023instruction,pyatkin2025generalizing,peng2025verif,qin2025incentivizing,guo2025ifdecorator}.
Broader evaluations further show that producing high-quality content does not ensure that a model satisfies multiple fine-grained user requirements \citep{jiang2023followbench,qin2024infobench}.
Most relevant to our setting, MathIF~\citep{fu2025scaling} and ReasonIF~\citep{kwon2025reasonif} evaluate instruction following in mathematical answers and reasoning traces.
Attention-based analyses further suggest that explicit reasoning can shift attention away from instruction-relevant tokens \citep{li2025thinkingfails}.
Together, these studies show that stronger reasoning does not guarantee compliance with constraints on format, language, length, or the reasoning trace.
Existing work mainly documents this gap at evaluation time.
We instead ask a training-time question: does optimizing one verifier preserve the rewardable support needed by the other objective?
Beyond math accuracy and instruction-following pass rate, we ask whether such training preserves the joint behavior users need.

\noindent\textbf{RLVR Mechanisms, Sequential Post-training, and Support Preservation.}
Recent work examines how RLVR changes model output distributions.
Token-level studies find that RLVR may change relatively few token probabilities yet substantially affect model behavior \citep{meng2026sparsecritical}.
Other studies ask whether RLVR learns new reasoning abilities or mainly reweights reasoning paths already present in the base policy \citep{yue2025does,wen2025rlvr}.
Work on continual learning, capability loss after alignment, and reward hacking shows that sequential optimization can cause forgetting or conflicts between objectives and can encourage models to exploit imperfect rewards \citep{kirkpatrick2017overcoming,ouyang2022training,amodei2016concrete,wen2024mislead}.
On-policy distillation (OPD) methods aim to stabilize post-training by using a teacher distribution to supervise the student's own rollouts \citep{song2026survey,wu2026lightningopd,xing2026trustregionopd}.
Our work studies a different limitation of on-policy training: after one verifier shifts the sampling distribution, rewardable trajectories for a later objective may become too rare to sample and reinforce, even when the relevant ability has not been broadly forgotten.
Sequential training and distillation should therefore be evaluated not only by the current verifier score, but also by whether they preserve rewardable support for future objectives.

\section{Experimental Design}

The core study first compares Math-RLVR and IF-RLVR from the same starting policy, then evaluates IF$\rightarrow$Math and Math$\rightarrow$IF to test whether the first stage affects later learning.

\paragraph{Training Paths.}
We use Qwen3-8B-Base and Qwen2.5-Math-7B, and denote each checkpoint before on-policy RLVR as Base.
Within each model family, both branches use the same rollout and evaluation protocol.
Math-RLVR trains on the 7.5k MATH split~\citep{hendrycks2021math} with an exact final-answer verifier, whereas IF-RLVR trains on IFTrain~\citep{pyatkin2025generalizing} with deterministic checkers that require all explicit response constraints to pass.
Each training configuration has one fixed-seed run, so repeated rollouts measure within-policy sampling variation rather than variation across training runs.
Complete training settings and mitigation recipes are provided in \Cref{app:experimental_setup}.

\paragraph{Evaluation and Support Measures.}
We evaluate math support on AIME24, AIME25, and MATH-500-128~\citep{hendrycks2021math,lightman2023lets}, and instruction-following support on IFEval~\citep{zhou2023instruction} and IFBench~\citep{pyatkin2025generalizing}.
MathIF~\citep{fu2025scaling} and ReasonIF~\citep{kwon2025reasonif} use a separate endpoint set for a joint-behavior stress test of whether correctness and constraint following occur in the same response.
For each prompt, we sample \(k\) stochastic rollouts and score them with the relevant verifier.
The rollout mean, reported as \(\metricat{mean}{k}\), estimates \(\metricat{pass}{1}\); \(\metricat{best}{k}\) records whether any rollout succeeds; and pass-count records the number of successes.
Core support analyses use \(k=32\) and group prompts as all-wrong, mixed, or all-correct when 0, 1--31, or 32 rollouts pass.
Benchmark roles, decoding settings, and joint-test filters are detailed in \Cref{app:benchmarks_and_protocols,app:external_joint_validation,app:if_data_lineage}.

\section{Verifier-Induced Cross-Task Support Reshaping}

Verifier choice changes not only current-task scores, but also which successful responses for the other task remain easy enough to sample.
Single-stage training exposes two asymmetric shifts: Math-RLVR raises average IF success while leaving fewer prompts solvable under repeated sampling, whereas IF-RLVR makes correct math responses harder to reach.
Sequential training then shows the consequence: IF-first training reduces the reward variation available to later Math-RLVR, while Math-first training requires a trade-off between math retention and IF gains.

\subsection{Math-RLVR Polarizes IF Support, Whereas IF-RLVR Reduces Math Searchability}
\label{sec:single_stage}

\subsubsection{Math-RLVR Raises Average IF Success but Reduces Prompt Coverage}
\label{sec:if_metric_divergence_and_polarization}

\begin{paperwraptable}{0.58\textwidth}
  \normalsize
  \setlength{\tabcolsep}{5pt}
  \renewcommand{\arraystretch}{1.10}
  \begin{tabularx}{\linewidth}{@{}l*{4}{>{\centering\arraybackslash}X}@{}}
    \toprule
    \rowcolor{white}
    & \multicolumn{2}{c}{Qwen3-8B-Base}
    & \multicolumn{2}{c}{Qwen2.5-Math-7B} \\
    \cmidrule(lr){2-3} \cmidrule(l){4-5}
    \rowcolor{white}
    Dataset
    & $\Delta\metricat{pass}{1}$
    & $\Delta\metricat{best}{32}$
    & $\Delta\metricat{pass}{1}$
    & $\Delta\metricat{best}{32}$ \\
    \midrule
    IFEval
    & +0.065 & -0.098
    & +0.079 & -0.114 \\
    \rowcolor{reportTableHighlight}
    IFBench
    & +0.032 & -0.067
    & +0.016 & -0.037 \\
    \bottomrule
  \end{tabularx}
  \caption{
  \textbf{Math-RLVR raises IF \(\metricat{pass}{1}\) but lowers \(\metricat{best}{32}\) across models and benchmarks.}
  Values report final-minus-Base changes.
  }
  \label{tab:if_delta}
\end{paperwraptable}

\begin{figure}[tbp]
  \centering
  \papergraphic{\linewidth}{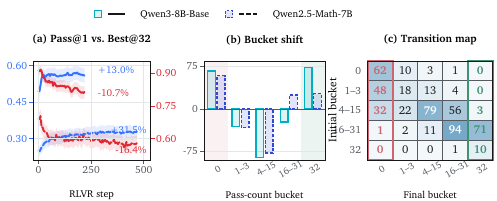}
  \caption{
  \textbf{Math-RLVR raises average IF success while polarizing prompt outcomes.}
  \textbf{(a)} \(\metricat{pass}{1}\) rises, whereas \(\metricat{best}{32}\) falls.
  \textbf{(b,c)} Mixed prompts move toward the all-correct and all-wrong buckets.
  }
  \label{fig:if_support_shift}
\end{figure}

Math-RLVR improves average IF success but reduces the number of prompts with any successful rollout under repeated sampling.
Across both model families and both IF benchmarks, \(\metricat{pass}{1}\) increases while \(\metricat{best}{32}\) decreases (\Cref{fig:if_support_shift}a; \Cref{tab:if_delta}).
For example, on IFEval with Qwen3-8B-Base, \(\metricat{pass}{1}\) rises by \(0.065\), whereas \(\metricat{best}{32}\) falls by \(0.098\).
The prompt-level analysis explains this gap: prompts in the mixed buckets move toward both the all-correct and all-wrong extremes (\Cref{fig:if_support_shift}b,c).
The policy therefore succeeds more often on an average rollout, but additional sampling recovers fewer prompts.

\subsubsection{IF-RLVR Lowers Math Searchability as Openings Shift}
\label{sec:routing_collapse}

\begin{figure}[tbp]
  \centering
  \papergraphic{\linewidth}{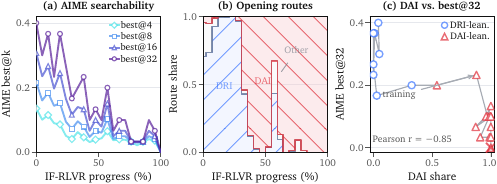}
  \caption{
  \textbf{IF-RLVR lowers math searchability as response openings shift from DRI to DAI.}
  \textbf{(a)} \(\metricat{best}{k}\) falls at every sampling budget; \textbf{(b)} DAI eventually dominates; \textbf{(c)} higher DAI share co-occurs with lower \(\metricat{best}{32}\).
  }
  \label{fig:if_rlvr_opening_routes}
\end{figure}

IF-RLVR steadily reduces math searchability while changing how sampled responses begin.
Across IF-RLVR checkpoints, AIME \(\metricat{best}{k}\) declines for every tested budget from \(k=4\) to \(k=32\) (\Cref{fig:if_rlvr_opening_routes}a).
To describe what changes in the sampled responses, we classify each rollout by its visible opening.
Deliberative reasoning initiation (DRI) begins with an explicit step-by-step derivation, direct answer initiation (DAI) exposes an answer before such a derivation, and unmatched openings are labeled Other.
These labels describe only the visible opening, not the model's internal reasoning state; the complete rules are provided in \Cref{app:route_classifier}.

As IF-RLVR progresses, DRI openings decline and DAI openings eventually dominate (\Cref{fig:if_rlvr_opening_routes}b).
Across checkpoints, higher DAI share co-occurs with lower AIME \(\metricat{best}{32}\) (\Cref{fig:if_rlvr_opening_routes}c).
The evidence in this section remains behavioral, so it does not show that mathematical ability has been erased or that DAI causes lower searchability.
We test the latter question with distributional analyses and controlled interventions in \Cref{sec:mechanism}.

\subsection{IF-First Training Leaves Little Reward Variation for Math-RLVR}
\label{sec:starvation}

IF-first training directly tests whether the single-stage support shift limits later on-policy learning.
Under the group-relative update used here, only prompts with both successful and failed rollouts provide within-group reward variation (\Cref{app:standalone_support_trajectories}).
The mixed group therefore tracks the prompts that can still provide a learning signal for later training.
We continue Math-RLVR from the IF-RLVR endpoint and compare this sequential path with the two standalone trajectories.

\begin{paperwrapfigure}{0.52\textwidth}
  \centering
  \papergraphic{\linewidth}{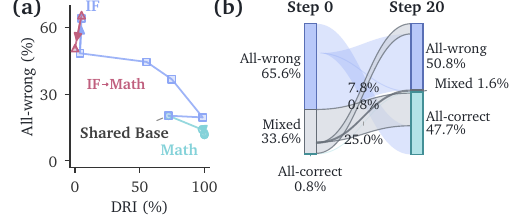}
  \caption{
  \textbf{After IF-first training, Math-RLVR improves outcomes without restoring DRI or mixed support.}
  \textbf{(a)} The all-wrong share falls, but DRI remains below standalone Math-RLVR; \textbf{(b)} the mixed group falls from 33.6\% to 1.6\% by step 20.
  }
  \label{fig:if2math_support_trajectory}
\end{paperwrapfigure}

After the switch to Math-RLVR, the all-wrong share falls from 65.6\% to 50.8\%, but DRI does not recover over the observed 20 steps (\Cref{fig:if2math_support_trajectory}a).
Over the same interval, the mixed group falls from 33.6\% to 1.6\%, while the all-correct group rises to 47.7\% (\Cref{fig:if2math_support_trajectory}b).
By step 20, almost every prompt lies in an all-wrong or all-correct group, leaving little within-group reward variation for subsequent updates.
Later Math-RLVR therefore improves current outcomes without rebuilding the DRI and mixed support observed under standalone Math-RLVR.
This conclusion applies only to the observed training path and does not imply an irreversible loss of mathematical reasoning ability.

\subsection{Math-First Training Trades IF Gains against Math Retention}
\label{sec:retention_adaptation}

\begin{figure}[tbp]
  \centering
  \papergraphic{\linewidth}{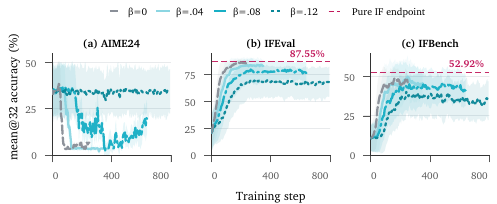}
  \caption{
  \textbf{Reference KL trades IF adaptation for math retention during Math$\rightarrow$IF training.}
  Curves and bands show \(\metricat{mean}{32}\) and \(\metricat{std}{32}\); dashed lines mark the standalone IF-RLVR endpoints.
  No tested coefficient retains math while matching the unconstrained IF gains.
  }
  \label{fig:math2if_kl_retention_adaptation}
\end{figure}

The reverse order exposes a different problem: unconstrained IF-RLVR improves IF performance but sharply reduces the math performance gained in the first stage.
When training continues from the Math-RLVR endpoint, \(\metricat{pass}{1}\) rises on both IF benchmarks but drops sharply on the math task (\Cref{fig:math2if_kl_retention_adaptation}).
To test whether a reference-policy constraint can retain math performance, we sweep the KL coefficient \(\beta\in\{0,0.04,0.08,0.12\}\) in Group Relative Policy Optimization (GRPO)~\citep{shao2024deepseekmath} during the second stage.

Weak KL constraints only delay the decline in math performance.
The \(\beta=0.12\) setting largely retains math-task \(\metricat{pass}{1}\), but it yields the smallest IF gains.
No tested coefficient therefore preserves math performance while matching the IF gains of the unconstrained baseline.

\begin{keytakeaway}
In the tested settings, neither training order provides a simple sequential solution: IF-first training reduces the mixed support needed by later math training, whereas Math-first training trades IF adaptation against math retention.
\end{keytakeaway}
The next section localizes the main distributional change induced by IF-RLVR and tests whether the response opening causally changes math searchability.

\section{Mechanism: Support Reshaping Concentrates at the Response Opening}
\label{sec:mechanism}

The previous section leaves two explanations for lower math searchability: IF-RLVR may broadly alter downstream reasoning, or it may mainly change which route the response enters.
Our evidence supports the second account.
Differences between Base and RLVR peak at the first generated token, where Math-RLVR and IF-RLVR favor opposite routes.
Inference-time interventions further show that this opening choice changes subsequent math searchability.

\subsection{Distributional Shifts Peak at the First Token}
\label{sec:opening_localization}

We localize distributional changes by evaluating Base, Math-RLVR, and IF-RLVR with teacher forcing on matched prefixes from Base-sampled traces and aggregating Jensen--Shannon (JS) divergence by response position.
The analysis covers both starting policies and AIME, IFEval, and IFBench; \Cref{app:topk_divergence} gives the top-\(K\) approximation and full definition.

\begin{figure}[H]
  \centering
  \papergraphic{0.676\linewidth}{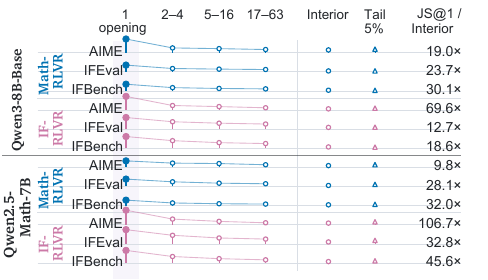}
  \caption{
  \textbf{RLVR changes next-token distributions most at the response opening.}
  Stems report mean JS over matched Base prefixes by position group, and right-hand values report the ratio of first-token JS to interior JS; absolute values and construction details are in \Cref{app:topk_divergence,tab:js-position-values}.
  }
  \label{fig:js_opening_localization}
\end{figure}

The first generated token has the highest mean JS for every combination of model, RLVR branch, and benchmark (\Cref{fig:js_opening_localization}).
On AIME, the ratio of first-token JS to interior JS ranges from \(9.8\times\) to \(106.7\times\) across the four combinations of model and RLVR branch.
Later positions show much smaller shifts, localizing the largest policy change to route entry rather than the full generation trace.

\subsection{RLVR Reranks Candidates toward Different Opening Routes}
\label{sec:candidate_reranking}
\label{sec:opening_route_direction}

Position-wise JS locates the shift but does not distinguish reranking existing candidates from promoting an opening that Base assigns little probability.
We therefore measure top-10 overlap and Base top-3 reuse at the highest-JS \(10\%\) of matched positions, and separately inspect the Base rank and probability of the RLVR top choice at AIME position 1.

\begin{table}[htbp]
\centering
\normalsize
\setlength{\tabcolsep}{2pt}
\renewcommand{\arraystretch}{1.10}
\rowcolors{3}{reportTableStripe}{white}
\begin{tabularx}{\linewidth}{@{}>{\raggedright\arraybackslash}p{0.20\linewidth}>{\raggedright\arraybackslash}p{0.08\linewidth}*{8}{>{\raggedleft\arraybackslash}X}@{}}
\toprule
\rowcolor{white}
& & \multicolumn{2}{c}{AIME} & \multicolumn{2}{c}{IFEval} & \multicolumn{2}{c}{IFBench} & \multicolumn{2}{c}{AIME position 1} \\
\cmidrule(lr){3-4}\cmidrule(lr){5-6}\cmidrule(lr){7-8}\cmidrule(lr){9-10}
\rowcolor{white}
Starting policy & RLVR tasks & Shared & \makecell{Top-3\\(\%)} & Shared & \makecell{Top-3\\(\%)} & Shared & \makecell{Top-3\\(\%)} & Rank & Prob. (\%) \\
\midrule
Qwen3-8B-Base & Math & 6.88 & 91.8 & 8.92 & 98.6 & 8.94 & 99.4 & 6 & 4.5 \\
Qwen3-8B-Base & IF & 8.51 & 98.3 & 7.40 & 90.0 & 7.72 & 96.0 & 14.5 & 1.9 \\
Qwen2.5-Math-7B & Math & 8.14 & 96.3 & 8.46 & 97.3 & 8.96 & 99.3 & 1 & 39.4 \\
Qwen2.5-Math-7B & IF & 7.80 & 94.3 & 7.06 & 89.3 & 8.00 & 94.7 & 18 & 0.4 \\
\bottomrule
\end{tabularx}
\caption{
\textbf{RLVR mostly reranks Base-supported candidates but can promote low-probability opening tokens.}
At the highest-JS \(10\%\) of matched-prefix positions, Shared/Top-3 report Base--RLVR top-10 overlap and Base top-3 reuse; at AIME position 1, Rank/Prob. report the Base rank and top-64-renormalized probability of the RLVR top choice.
}
\label{tab:high-js-candidate-reuse}
\end{table}

At high-JS positions, Base and RLVR share \(6.88\) to \(8.96\) top-10 candidates on average, and \(89.3\%\) to \(99.4\%\) of RLVR top choices already appear in the Base top-3 (\Cref{tab:high-js-candidate-reuse}).
At the AIME opening under IF-RLVR, however, the top choice has median Base ranks of \(14.5\) and \(18\) across the two starting policies, with mean probabilities of only \(1.9\%\) and \(0.4\%\).
RLVR therefore usually reweights continuations already supported by Base, but it can also make a low-probability first token dominant.

To determine the direction of this reranking, we compare first-token probabilities on the same AIME prompts and group candidates as DRI-like, neutral, or DAI-like according to their visible openings.
These labels describe output form rather than the model's internal reasoning state.

\begin{figure}[H]
  \centering
  \papergraphic{0.855\linewidth}{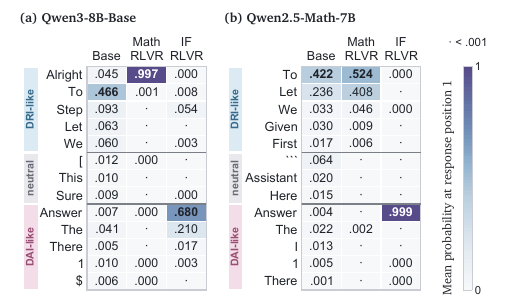}
  \caption{
  \textbf{Verifier choice shifts first-token probability toward different response routes.}
  Each cell reports the mean probability at AIME response position 1; rows group tokens by the visible openings they initiate.
  }
  \label{fig:first_token_route_direction}
\end{figure}

Across both models, the top-ranked first token under Math-RLVR is DRI-like (``Alright'' or ``To''), whereas IF-RLVR makes the DAI-like token ``Answer'' the top choice (\Cref{fig:first_token_route_direction}).
Thus, the two verifiers push generation toward opposite routes at the first token.

\subsection{Changing the Opening Route Changes Math Searchability}
\label{sec:opening_route_causal_bottleneck}

The distributional results above do not establish whether the opening route itself changes downstream math search.
We therefore intervene without updating model parameters: under Base decoding, we force an IF-side token or DAI prefix; under IF-RLVR decoding, we force a Base-side token or DRI prefix and compare against free decoding.
We also move the same IF-side token across decoded positions to test whether its route effect is specific to the response opening; full definitions are in \Cref{app:opening_interventions}.

\begin{figure}[H]
  \centering
  \papergraphic{0.82\linewidth}{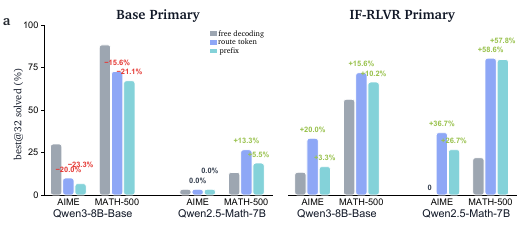}
  \par\medskip
  \papergraphic{0.65\linewidth}{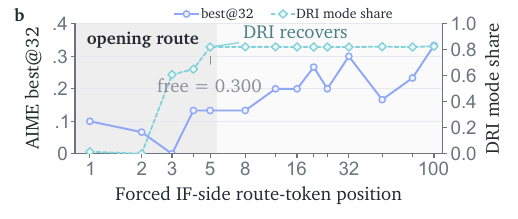}
  \caption{
  \textbf{Changing the response opening changes math searchability.}
  \textbf{(a)} \(\metricat{best}{32}\) under free decoding, a forced route token, and a forced prefix; labels report changes (\%) from free-decoding baselines.
  \textbf{(b)} \(\metricat{best}{32}\) and DRI share when the same IF-side token is forced at different positions; shading marks the response opening.
  }
  \label{fig:opening_route_causal_bottleneck}
\end{figure}

From IF-RLVR checkpoints, forcing Base-side or DRI openings raises \(\metricat{best}{32}\) for both model families on AIME and MATH-500 (\Cref{fig:opening_route_causal_bottleneck}a).
The reverse intervention is not fully symmetric: IF-side or DAI openings reduce searchability for Qwen3-8B-Base but have neutral or positive effects for Qwen2.5-Math-7B.
In the Qwen3-8B-Base AIME position sweep, the same IF-side token suppresses the DRI share only near the response opening; the DRI route recovers at later positions, although \(\metricat{best}{32}\) remains variable (\Cref{fig:opening_route_causal_bottleneck}b).
\begin{keytakeaway}
Together, these results identify a causal role for opening-route selection in math searchability within the tested settings.
They support a localized change in route entry rather than a broad rewrite of downstream reasoning.
\end{keytakeaway}
This mechanism motivates the routing-prior and distillation experiments in \Cref{sec:mitigation_levers_limits}.

\section{Mitigating Support Reshaping: Levers and Limits}
\label{sec:mitigation_levers_limits}

Within the tested settings, \Cref{sec:mechanism} shows that the response opening causally affects math searchability.
We therefore test two mitigation strategies: a DRI-biased SFT prior before IF-RLVR and on-policy distillation (OPD) from token-level teacher distributions.
The first only delays DAI dominance, and the second depends on the teacher state; neither consistently preserves math support.

\subsection{A DRI Prior Delays but Does Not Prevent DAI Dominance}
\label{sec:routing_prior_cold_start}

We construct four SFT cold starts from 50 correct math responses to test whether an initial route prior survives subsequent IF-RLVR.
Soft DRI and Hard DRI use the same DRI-only corpus with different SFT strengths, whereas the DAI and Random controls use DAI-only and mixed openings under the stronger configuration.
All four policies then receive the same 100-step IF-RLVR training; complete settings are provided in \Cref{app:routing_prior_cold_start}.

\begin{figure}[tbp]
  \centering
  \papergraphic{0.816\linewidth}{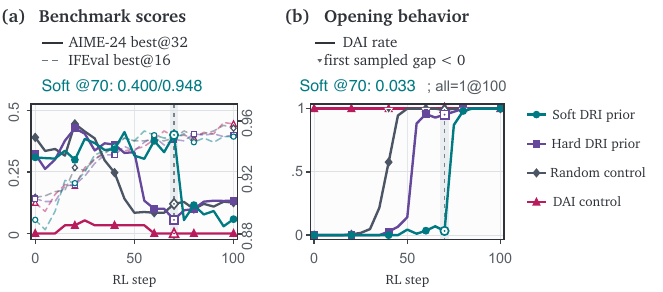}
  \caption{
  \textbf{A DRI-biased prior delays but does not prevent DAI dominance.}
  \textbf{(a)} Solid lines show AIME24 \(\metricat{best}{32}\), and dashed lines show IFEval \(\metricat{best}{16}\); \textbf{(b)} curves show the AIME24 DAI rate.
  }
  \label{fig:routing_prior_cold_start}
\end{figure}

Soft DRI delays the shift longest: at step 70, it retains AIME24 \(\metricat{best}{32}=0.400\) and IFEval \(\metricat{best}{16}=0.948\), with a DAI rate of only \(0.033\).
By step 100, however, every condition reaches a DAI rate of 1, and the Hard DRI prior does not last longer than the weaker Soft DRI prior in this run.
A one-time route prior therefore changes when the shift occurs but not the observed endpoint route.

\subsection{OPD Outcomes Depend on the Teacher State}

We next test whether dense supervision preserves support more reliably than a one-time prior: a Base-initialized OPD student matches an IF-RLVR teacher's token distribution on student-generated rollouts.
Dense supervision alone is insufficient: whether the student retains math support depends on the selected teacher state.

\subsubsection{A Converged IF Teacher Does Not Preserve Math Support}
\label{sec:full_teacher_opd}

We first use the step-100 IF-RLVR endpoint as the teacher and train the OPD student for 50 steps.
MATH-500-128 \(\metricat{mean}{16}\) falls from \(0.3433\) to \(0.0879\), while the shortcut-response share rises from \(4.5\%\) to \(18.8\%\).

To examine which teacher preferences transfer, we compute the teacher and student shifts in token-averaged log probability relative to Base, denoted by \(\Delta_T\) and \(\Delta_S\), on the same 6,400 student-generated rollouts.
If the student directly reproduced teacher preferences, larger \(\Delta_T\) would coincide with larger \(\Delta_S\).

\begin{figure}[tbp]
  \centering
  \papergraphic{0.877\linewidth}{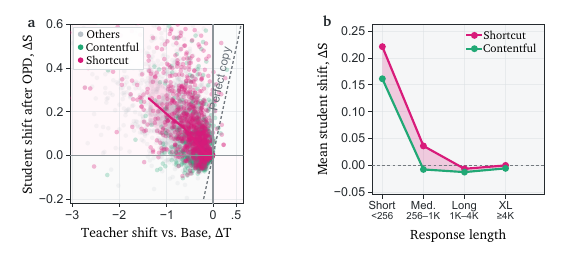}
  \captionsetup{skip=2pt}
  \caption{
  \textbf{OPD from a converged IF teacher does not reproduce teacher preferences rollout by rollout.}
  \textbf{(a)} The gray dashed line marks equal teacher and student shifts; \textbf{(b)} positive student shifts concentrate on short responses.
  }
  \label{fig:full_teacher_opd}
\end{figure}

Instead, teacher and student shifts have a Spearman correlation of \(\rho=-0.594\), and positive student shifts occur mainly on short responses.
Dense supervision from a converged teacher therefore does not automatically preserve cross-task support.

\subsubsection{Different Teacher States Favor Different Outcomes}
\label{sec:opd_teacher_state_scan}

To test earlier teachers, we select IF-RLVR snapshots from steps 20, 40, 60, and 80 and train each Base-initialized student for 100 OPD steps.
This coarse scan of four states cannot identify the best teacher along the full trajectory.

\begin{figure}[H]
  \centering
  \papergraphic{0.85\linewidth}{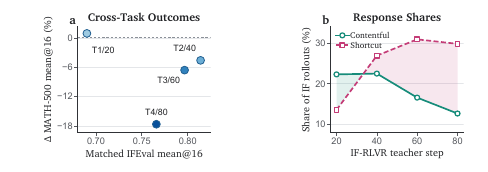}
  \captionsetup{skip=2pt}
  \caption{
  \textbf{Different teacher states favor different OPD outcomes.}
  \textbf{(a)} Final-student matched IFEval \(\metricat{mean}{16}\) versus the change in MATH-500 \(\metricat{mean}{16}\) from Base; \textbf{(b)} contentful and shortcut response shares.
  }
  \label{fig:opd_teacher_state_scan}
\end{figure}

Among the four sampled states, the strongest target-task transfer and the strongest math-support retention occur with different teachers.
T2/40 reaches the highest IFEval \(\metricat{mean}{16}\) of \(0.814\), whereas T1/20 is the only student that improves MATH-500 \(\metricat{mean}{16}\), by \(+0.93\%\).
T1/20 also produces more contentful than shortcut responses, whereas T2/40 through T4/80 show the reverse.
The teacher's IF endpoint alone is therefore insufficient for selecting a state that preserves math support.

\begin{keytakeaway}
Overall, a one-time DRI prior is eventually overwritten by continued IF-RLVR, while OPD still requires teacher selection that accounts for IF transfer, math support, and response quality.
In these experiments, neither method preserves math support throughout the full IF adaptation path.
\end{keytakeaway}
The next section turns to an independent joint-support test of whether math and instruction-following gains coexist in the same response.

\section{Marginal Gains Translate Only Partly into Joint Support}
\label{sec:joint_support_stress_test}

Across two joint-capability benchmarks, the marginal gains from Math-RLVR and IF-RLVR translate only partly into joint support within the same response.
The earlier sections measure math and instruction-following support separately, but success on the two metrics can occur in different rollouts.
We therefore use independent step-720 Qwen3-8B-Base, Math-RLVR, and IF-RLVR endpoints as a joint-behavior stress test, not a continuation of the earlier trajectories.

MathIF~\citep{fu2025scaling} constrains the full mathematical response, whereas ReasonIF~\citep{kwon2025reasonif} constrains the reasoning trace while isolating the final answer.
For each rollout, we measure correctness \(C\), benchmark-specific strict following \(F\), and joint support \(J=C\land F\), which requires both events in the same response.
Sampling, filtering, and scoring details are provided in \Cref{app:external_joint_validation}.

\begin{table}[htbp]
\centering
\normalsize
\setlength{\tabcolsep}{6pt}
\renewcommand{\arraystretch}{1.10}
\rowcolors{2}{reportTableStripe}{white}
\begin{tabularx}{0.78\linewidth}{@{}>{\hsize=1.05\hsize\raggedright\arraybackslash}X>{\hsize=1.30\hsize\raggedright\arraybackslash}X>{\hsize=0.70\hsize\raggedleft\arraybackslash}X>{\hsize=0.70\hsize\raggedleft\arraybackslash}X>{\hsize=1.25\hsize\raggedleft\arraybackslash}X@{}}
\toprule
\rowcolor{white}
Benchmark & Checkpoint & \(C\uparrow\) & \(F\uparrow\) & \(J=C\land F\uparrow\) \\
\midrule
\textit{MathIF} & Base      & 0.359 & 0.250 & 0.094 \\
  & Math-RLVR & \textbf{0.472} & 0.253 & 0.118 \\
  & IF-RLVR   & 0.377 & \textbf{0.431} & \textbf{0.148} \\
\midrule
\textit{ReasonIF} & Base      & 0.467 & 0.239 & 0.109 \\
  & Math-RLVR & \textbf{0.522} & 0.235 & 0.126 \\
  & IF-RLVR   & 0.438 & \textbf{0.391} & \textbf{0.171} \\
\bottomrule
\end{tabularx}
\caption{
\textbf{Marginal and joint support on MathIF and ReasonIF.}
Blocks average 16 rollouts over 290 prompts; \(C/F/J\) denote correctness, strict following, and same-rollout conjunction; filters are in \Cref{app:external_joint_validation}.
}
\label{tab:ch7_joint_support}
\end{table}

On MathIF and ReasonIF, Math-RLVR raises \(C\) by \(11.3\%/5.6\%\), while \(F\) remains nearly unchanged; \(J\) rises by only \(2.4\%/1.7\%\).
IF-RLVR shows the complementary pattern: \(F\) rises by \(18.1\%/15.3\%\), while \(C\) increases only slightly on MathIF and decreases on ReasonIF; \(J\) rises by \(5.5\%/6.2\%\).
In both training directions, the non-target metric changes far less than the verifier-aligned metric and sometimes declines, limiting the observed joint gain.
Endpoint improvements on one objective therefore do not automatically produce responses that are both correct and constraint-following.
Because these endpoints are independent of the controlled trajectories above, the experiment provides an external joint-behavior check of this endpoint implication rather than a continuation of the earlier learning curves.

\section{Conclusion}

Verifier choice in on-policy RLVR shapes current scores and future trainability.
Math-RLVR polarizes IF support; IF-RLVR favors direct answers and reduces math searchability.
Controlled interventions show a causal role for route entry; joint tests show gains combine only partly.
Effective support depends on rollout budget, so future work should estimate and preserve it throughout cross-verifier optimization.

\clearpage
\bibliography{references}

\begin{thebibliography}{36}
\providecommand{\natexlab}[1]{#1}
\providecommand{\url}[1]{\texttt{#1}}
\expandafter\ifx\csname urlstyle\endcsname\relax
  \providecommand{\doi}[1]{doi: #1}\else
  \providecommand{\doi}{doi: \begingroup \urlstyle{rm}\Url}\fi

\bibitem[Amodei et~al.(2016)Amodei, Olah, Steinhardt, Christiano, Schulman, and
  Man{\'e}]{amodei2016concrete}
Dario Amodei, Chris Olah, Jacob Steinhardt, Paul Christiano, John Schulman, and
  Dan Man{\'e}.
\newblock Concrete problems in {AI} safety, 2016.
\newblock URL \url{https://arxiv.org/abs/1606.06565}.

\bibitem[Chen et~al.(2026)Chen, Razin, Narasimhan, and Chen]{chen2025retaining}
Howard Chen, Noam Razin, Karthik Narasimhan, and Danqi Chen.
\newblock Retaining by doing: The role of on-policy data in mitigating
  forgetting.
\newblock In \emph{Proceedings of the 43rd International Conference on Machine
  Learning}, 2026.
\newblock URL \url{https://icml.cc/virtual/2026/poster/64375}.

\bibitem[Cobbe et~al.(2021)Cobbe, Kosaraju, Bavarian, Chen, Jun, Kaiser,
  Plappert, Tworek, Hilton, Nakano, Hesse, and Schulman]{cobbe2021training}
Karl Cobbe, Vineet Kosaraju, Mohammad Bavarian, Mark Chen, Heewoo Jun, Lukasz
  Kaiser, Matthias Plappert, Jerry Tworek, Jacob Hilton, Reiichiro Nakano,
  Christopher Hesse, and John Schulman.
\newblock Training verifiers to solve math word problems, 2021.
\newblock URL \url{https://arxiv.org/abs/2110.14168}.

\bibitem[Fu et~al.(2026)Fu, Li, Gu, Qu, and Cheng]{fu2025scaling}
Tingchen Fu, Yafu Li, Jiawei Gu, Xiaoye Qu, and Yu~Cheng.
\newblock Scaling reasoning, losing control: Evaluating instruction following
  in large reasoning models.
\newblock In \emph{Proceedings of the 64th Annual Meeting of the Association
  for Computational Linguistics (Volume 1: Long Papers)}, pp.\  40445--40463.
  Association for Computational Linguistics, 2026.
\newblock \doi{10.18653/v1/2026.acl-long.1878}.
\newblock URL \url{https://aclanthology.org/2026.acl-long.1878/}.

\bibitem[Guo et~al.(2025{\natexlab{a}})Guo, Yang, Zhang, Song, Wang, Zhu, Xu,
  Zhang, Ma, Bi, Zhang, Yu, Wu, Wu, Gou, Shao, Li, Gao, Liu, Xue, Wang, Wu,
  Feng, Lu, Zhao, Deng, Ruan, Dai, Chen, Ji, Li, Lin, Dai, Luo, Hao, Chen, Li,
  Zhang, Xu, Ding, Gao, Qu, Li, Guo, Li, Chen, Yuan, Tu, Qiu, Li, Cai, Ni,
  Liang, Chen, Dong, Hu, You, Gao, Guan, Huang, Yu, Wang, Zhang, Zhao, Wang,
  Zhang, Xu, Xia, Zhang, Zhang, Tang, Zhou, Li, Wang, Li, Tian, Huang, Zhang,
  Wang, Chen, Du, Ge, Zhang, Pan, Wang, Chen, Jin, Chen, Lu, Zhou, Chen, Ye,
  Wang, Yu, Zhou, Pan, Li, Zhou, Wu, Yun, Pei, Sun, Wang, Zeng, Liu, Liang,
  Gao, Yu, Zhang, Xiao, An, Liu, Wang, Chen, Nie, Cheng, Liu, Xie, Liu, Yang,
  Li, Su, Lin, Li, Jin, Shen, Chen, Sun, Wang, Song, Zhou, Wang, Shan, Li,
  Wang, Wei, Zhang, Xu, Li, Zhao, Sun, Wang, Yu, Zhang, Shi, Xiong, He, Piao,
  Wang, Tan, Ma, Liu, Guo, Ou, Wang, Gong, Zou, He, Xiong, Luo, You, Liu, Zhou,
  Zhu, Huang, Li, Zheng, Zhu, Ma, Tang, Zha, Yan, Ren, Ren, Sha, Fu, Xu, Xie,
  Zhang, Hao, Ma, Yan, Wu, Gu, Zhu, Liu, Li, Xie, Song, Pan, Huang, Xu, Zhang,
  and Zhang]{deepseekai2025deepseekr1}
Daya Guo, Dejian Yang, Haowei Zhang, Junxiao Song, Peiyi Wang, Qihao Zhu,
  Runxin Xu, Ruoyu Zhang, Shirong Ma, Xiao Bi, Xiaokang Zhang, Xingkai Yu,
  Yu~Wu, Z.~F. Wu, Zhibin Gou, Zhihong Shao, Zhuoshu Li, Ziyi Gao, Aixin Liu,
  Bing Xue, Bingxuan Wang, Bochao Wu, Bei Feng, Chengda Lu, Chenggang Zhao,
  Chengqi Deng, Chong Ruan, Damai Dai, Deli Chen, Dongjie Ji, Erhang Li,
  Fangyun Lin, Fucong Dai, Fuli Luo, Guangbo Hao, Guanting Chen, Guowei Li,
  H.~Zhang, Hanwei Xu, Honghui Ding, Huazuo Gao, Hui Qu, Hui Li, Jianzhong Guo,
  Jiashi Li, Jingchang Chen, Jingyang Yuan, Jinhao Tu, Junjie Qiu, Junlong Li,
  J.~L. Cai, Jiaqi Ni, Jian Liang, Jin Chen, Kai Dong, Kai Hu, Kaichao You,
  Kaige Gao, Kang Guan, Kexin Huang, Kuai Yu, Lean Wang, Lecong Zhang, Liang
  Zhao, Litong Wang, Liyue Zhang, Lei Xu, Leyi Xia, Mingchuan Zhang, Minghua
  Zhang, Minghui Tang, Mingxu Zhou, Meng Li, Miaojun Wang, Mingming Li, Ning
  Tian, Panpan Huang, Peng Zhang, Qiancheng Wang, Qinyu Chen, Qiushi Du, Ruiqi
  Ge, Ruisong Zhang, Ruizhe Pan, Runji Wang, R.~J. Chen, R.~L. Jin, Ruyi Chen,
  Shanghao Lu, Shangyan Zhou, Shanhuang Chen, Shengfeng Ye, Shiyu Wang,
  Shuiping Yu, Shunfeng Zhou, Shuting Pan, S.~S. Li, Shuang Zhou, Shaoqing Wu,
  Tao Yun, Tian Pei, Tianyu Sun, T.~Wang, Wangding Zeng, Wen Liu, Wenfeng
  Liang, Wenjun Gao, Wenqin Yu, Wentao Zhang, W.~L. Xiao, Wei An, Xiaodong Liu,
  Xiaohan Wang, Xiaokang Chen, Xiaotao Nie, Xin Cheng, Xin Liu, Xin Xie,
  Xingchao Liu, Xinyu Yang, Xinyuan Li, Xuecheng Su, Xuheng Lin, X.~Q. Li,
  Xiangyue Jin, Xiaojin Shen, Xiaosha Chen, Xiaowen Sun, Xiaoxiang Wang, Xinnan
  Song, Xinyi Zhou, Xianzu Wang, Xinxia Shan, Y.~K. Li, Y.~Q. Wang, Y.~X. Wei,
  Yang Zhang, Yanhong Xu, Yao Li, Yao Zhao, Yaofeng Sun, Yaohui Wang, Yi~Yu,
  Yichao Zhang, Yifan Shi, Yiliang Xiong, Ying He, Yishi Piao, Yisong Wang,
  Yixuan Tan, Yiyang Ma, Yiyuan Liu, Yongqiang Guo, Yuan Ou, Yuduan Wang, Yue
  Gong, Yuheng Zou, Yujia He, Yunfan Xiong, Yuxiang Luo, Yuxiang You, Yuxuan
  Liu, Yuyang Zhou, Y.~X. Zhu, Yanping Huang, Yaohui Li, Yi~Zheng, Yuchen Zhu,
  Yunxian Ma, Ying Tang, Yukun Zha, Yuting Yan, Z.~Z. Ren, Zehui Ren, Zhangli
  Sha, Zhe Fu, Zhean Xu, Zhenda Xie, Zhengyan Zhang, Zhewen Hao, Zhicheng Ma,
  Zhigang Yan, Zhiyu Wu, Zihui Gu, Zijia Zhu, Zijun Liu, Zilin Li, Ziwei Xie,
  Ziyang Song, Zizheng Pan, Zhen Huang, Zhipeng Xu, Zhongyu Zhang, and Zhen
  Zhang.
\newblock {DeepSeek-R1} incentivizes reasoning in {LLMs} through reinforcement
  learning.
\newblock \emph{Nature}, 645\penalty0 (8081):\penalty0 633--638,
  2025{\natexlab{a}}.
\newblock \doi{10.1038/s41586-025-09422-z}.
\newblock URL \url{https://doi.org/10.1038/s41586-025-09422-z}.

\bibitem[Guo et~al.(2025{\natexlab{b}})Guo, Liang, Jian, Yang, Wu, Li, Lu, Guo,
  and Chen]{guo2025ifdecorator}
Xu~Guo, Tianyi Liang, Tong Jian, Xiaogui Yang, Ling-I Wu, Chenhui Li, Zhihui
  Lu, Qipeng Guo, and Kai Chen.
\newblock {IFDECORATOR}: Wrapping instruction following reinforcement learning
  with verifiable rewards, 2025{\natexlab{b}}.
\newblock URL \url{https://arxiv.org/abs/2508.04632}.

\bibitem[Hendrycks et~al.(2021)Hendrycks, Burns, Kadavath, Arora, Basart, Tang,
  Song, and Steinhardt]{hendrycks2021math}
Dan Hendrycks, Collin Burns, Saurav Kadavath, Akul Arora, Steven Basart, Eric
  Tang, Dawn Song, and Jacob Steinhardt.
\newblock Measuring mathematical problem solving with the {MATH} dataset.
\newblock In \emph{Proceedings of the Neural Information Processing Systems
  Track on Datasets and Benchmarks}, 2021.
\newblock URL
  \url{https://datasets-benchmarks-proceedings.neurips.cc/paper/2021/hash/be83ab3ecd0db773eb2dc1b0a17836a1-Abstract-round2.html}.

\bibitem[Jiang et~al.(2024)Jiang, Wang, Zeng, Zhong, Li, Mi, Shang, Jiang, Liu,
  and Wang]{jiang2023followbench}
Yuxin Jiang, Yufei Wang, Xingshan Zeng, Wanjun Zhong, Liangyou Li, Fei Mi,
  Lifeng Shang, Xin Jiang, Qun Liu, and Wei Wang.
\newblock {FollowBench}: A multi-level fine-grained constraints following
  benchmark for large language models.
\newblock In \emph{Proceedings of the 62nd Annual Meeting of the Association
  for Computational Linguistics (Volume 1: Long Papers)}, pp.\  4667--4688.
  Association for Computational Linguistics, 2024.
\newblock \doi{10.18653/v1/2024.acl-long.257}.
\newblock URL \url{https://aclanthology.org/2024.acl-long.257/}.

\bibitem[{Kimi Team} et~al.(2025){Kimi Team}, Du, Gao, Xing, Jiang, Chen, Li,
  Xiao, Du, Liao, Tang, Wang, Zhang, Yuan, Lu, Tang, Sung, Wei, Lai, Guo, Zhu,
  Ding, Hu, Yang, Zhang, Yao, Zhao, Lu, Li, Yu, Gao, Zheng, Yuan, Chen, Guo,
  Su, Wang, Zhao, Zhang, Liu, Yan, Wu, Shi, Ye, Yu, Dong, Zhang, Ma, Pan, Gong,
  Liu, Ma, Wei, Cao, Huang, Jiang, Gao, Xiong, He, Huang, Xu, Wu, He, Wei, Jia,
  Wu, Xu, Zu, Zhou, Pan, Charles, Li, Hu, Liu, Chen, Wang, Liu, Qin, Liu, Yang,
  Bao, Du, Wu, Wang, Zhou, Wang, Li, Zhu, Zhang, Wang, Yang, Huang, Huang, Xu,
  Yang, and Lin]{kimiteam2025kimi}
{Kimi Team}, Angang Du, Bofei Gao, Bowei Xing, Changjiu Jiang, Cheng Chen,
  Cheng Li, Chenjun Xiao, Chenzhuang Du, Chonghua Liao, Chuning Tang, Congcong
  Wang, Dehao Zhang, Enming Yuan, Enzhe Lu, Fengxiang Tang, Flood Sung, Guangda
  Wei, Guokun Lai, Haiqing Guo, Han Zhu, Hao Ding, Hao Hu, Hao Yang, Hao Zhang,
  Haotian Yao, Haotian Zhao, Haoyu Lu, Haoze Li, Haozhen Yu, Hongcheng Gao,
  Huabin Zheng, Huan Yuan, Jia Chen, Jianhang Guo, Jianlin Su, Jianzhou Wang,
  Jie Zhao, Jin Zhang, Jingyuan Liu, Junjie Yan, Junyan Wu, Lidong Shi, Ling
  Ye, Longhui Yu, Mengnan Dong, Neo Zhang, Ningchen Ma, Qiwei Pan, Qucheng
  Gong, Shaowei Liu, Shengling Ma, Shupeng Wei, Sihan Cao, Siying Huang, Tao
  Jiang, Weihao Gao, Weimin Xiong, Weiran He, Weixiao Huang, Weixin Xu, Wenhao
  Wu, Wenyang He, Xianghui Wei, Xianqing Jia, Xingzhe Wu, Xinran Xu, Xinxing
  Zu, Xinyu Zhou, Xuehai Pan, Y.~Charles, Yang Li, Yangyang Hu, Yangyang Liu,
  Yanru Chen, Yejie Wang, Yibo Liu, Yidao Qin, Yifeng Liu, Ying Yang, Yiping
  Bao, Yulun Du, Yuxin Wu, Yuzhi Wang, Zaida Zhou, Zhaoji Wang, Zhaowei Li,
  Zhen Zhu, Zheng Zhang, Zhexu Wang, Zhilin Yang, Zhiqi Huang, Zihao Huang,
  Ziyao Xu, Zonghan Yang, and Zongyu Lin.
\newblock {Kimi k1.5}: Scaling reinforcement learning with {LLMs}, 2025.
\newblock URL \url{https://arxiv.org/abs/2501.12599}.

\bibitem[Kirkpatrick et~al.(2017)Kirkpatrick, Pascanu, Rabinowitz, Veness,
  Desjardins, Rusu, Milan, Quan, Ramalho, Grabska-Barwinska, Hassabis, Clopath,
  Kumaran, and Hadsell]{kirkpatrick2017overcoming}
James Kirkpatrick, Razvan Pascanu, Neil Rabinowitz, Joel Veness, Guillaume
  Desjardins, Andrei~A. Rusu, Kieran Milan, John Quan, Tiago Ramalho, Agnieszka
  Grabska-Barwinska, Demis Hassabis, Claudia Clopath, Dharshan Kumaran, and
  Raia Hadsell.
\newblock Overcoming catastrophic forgetting in neural networks.
\newblock \emph{Proceedings of the National Academy of Sciences}, 114\penalty0
  (13):\penalty0 3521--3526, 2017.
\newblock \doi{10.1073/pnas.1611835114}.
\newblock URL \url{https://doi.org/10.1073/pnas.1611835114}.

\bibitem[Kwon et~al.(2026)Kwon, Zhu, Bianchi, Zhou, and Zou]{kwon2025reasonif}
Yongchan Kwon, Shang Zhu, Federico Bianchi, Kaitlyn Zhou, and James Zou.
\newblock {ReasonIF}: Large reasoning models fail to follow instructions during
  reasoning.
\newblock In \emph{Findings of the Association for Computational Linguistics:
  {ACL} 2026}, pp.\  29149--29164. Association for Computational Linguistics,
  2026.
\newblock \doi{10.18653/v1/2026.findings-acl.1456}.
\newblock URL \url{https://aclanthology.org/2026.findings-acl.1456/}.

\bibitem[Lai et~al.(2026)Lai, Zhao, Feng, Ma, Liu, Zhao, Lin, Yi, Zhang, Liu,
  Meng, and Zhu]{lai2025rftforgetting}
Song Lai, Haohan Zhao, Rong Feng, Changyi Ma, Wenzhuo Liu, Hongbo Zhao, Xi~Lin,
  Dong Yi, Qingfu Zhang, Hongbin Liu, Gaofeng Meng, and Fei Zhu.
\newblock Reinforcement fine-tuning naturally mitigates forgetting in continual
  post-training, 2026.
\newblock URL \url{https://arxiv.org/abs/2507.05386}.

\bibitem[Li et~al.(2026)Li, Zhang, Wei, Gao, Guo, Luo, Song, Huang, and
  Wang]{li2026sweext}
Wei Li, Xin Zhang, Shaohang Wei, Yanjie Gao, Zhongxin Guo, Wen Luo, Feifan
  Song, Yangyu Huang, and Houfeng Wang.
\newblock {{SWE}-Ext}: Extending and scaling augmented data for
  repository-level coding tasks, 2026.
\newblock URL \url{https://openreview.net/forum?id=HYQXJYzmFU}.

\bibitem[Li et~al.(2025)Li, Yu, Zhang, Chen, Zhang, Zhuang, Sadagopan, and
  Beniwal]{li2025thinkingfails}
Xiaomin Li, Zhou Yu, Zhiwei Zhang, Xupeng Chen, Ziji Zhang, Yingying Zhuang,
  Narayanan Sadagopan, and Anurag Beniwal.
\newblock When thinking fails: The pitfalls of reasoning for
  instruction-following in {LLMs}.
\newblock In \emph{Advances in Neural Information Processing Systems},
  volume~38, pp.\  77925--77962. Curran Associates, Inc., 2025.
\newblock URL
  \url{https://proceedings.neurips.cc/paper_files/paper/2025/hash/706338a08f9378b708f21cbf5686e617-Abstract-Conference.html}.

\bibitem[Lightman et~al.(2024)Lightman, Kosaraju, Burda, Edwards, Baker, Lee,
  Leike, Schulman, Sutskever, and Cobbe]{lightman2023lets}
Hunter Lightman, Vineet Kosaraju, Yura Burda, Harri Edwards, Bowen Baker, Teddy
  Lee, Jan Leike, John Schulman, Ilya Sutskever, and Karl Cobbe.
\newblock Let's verify step by step.
\newblock In \emph{International Conference on Learning Representations}, 2024.
\newblock URL \url{https://openreview.net/forum?id=v8L0pN6EOi}.

\bibitem[Meng et~al.(2026)Meng, Huang, Wei, Ma, Yang, Wang, Wang, Ding, and
  Zhou]{meng2026sparsecritical}
Haoming Meng, Kexin Huang, Shaohang Wei, Chiyu Ma, Shuo Yang, Xue Wang, Guoyin
  Wang, Bolin Ding, and Jingren Zhou.
\newblock Sparse but critical: A token-level analysis of distributional shifts
  in {RLVR} fine-tuning of {LLMs}.
\newblock In \emph{International Conference on Learning Representations}, 2026.
\newblock URL \url{https://openreview.net/forum?id=8vWIXno8LW}.

\bibitem[Ouyang et~al.(2022)Ouyang, Wu, Jiang, Almeida, Wainwright, Mishkin,
  Zhang, Agarwal, Slama, Ray, Schulman, Hilton, Kelton, Miller, Simens, Askell,
  Welinder, Christiano, Leike, and Lowe]{ouyang2022training}
Long Ouyang, Jeffrey Wu, Xu~Jiang, Diogo Almeida, Carroll Wainwright, Pamela
  Mishkin, Chong Zhang, Sandhini Agarwal, Katarina Slama, Alex Ray, John
  Schulman, Jacob Hilton, Fraser Kelton, Luke Miller, Maddie Simens, Amanda
  Askell, Peter Welinder, Paul~F. Christiano, Jan Leike, and Ryan Lowe.
\newblock Training language models to follow instructions with human feedback.
\newblock In \emph{Advances in Neural Information Processing Systems},
  volume~35, 2022.
\newblock URL
  \url{https://proceedings.neurips.cc/paper_files/paper/2022/hash/b1efde53be364a73914f58805a001731-Abstract-Conference.html}.

\bibitem[Parisi et~al.(2019)Parisi, Kemker, Part, Kanan, and
  Wermter]{parisi2019continual}
German~I. Parisi, Ronald Kemker, Jose~L. Part, Christopher Kanan, and Stefan
  Wermter.
\newblock Continual lifelong learning with neural networks: A review.
\newblock \emph{Neural Networks}, 113:\penalty0 54--71, 2019.
\newblock \doi{10.1016/j.neunet.2019.01.012}.
\newblock URL \url{https://arxiv.org/abs/1802.07569}.

\bibitem[Peng et~al.(2025)Peng, Qi, Wang, Xu, Hou, and Li]{peng2025verif}
Hao Peng, Yunjia Qi, Xiaozhi Wang, Bin Xu, Lei Hou, and Juanzi Li.
\newblock {VerIF}: Verification engineering for reinforcement learning in
  instruction following.
\newblock In \emph{Proceedings of the 2025 Conference on Empirical Methods in
  Natural Language Processing}, pp.\  30324--30339. Association for
  Computational Linguistics, 2025.
\newblock \doi{10.18653/v1/2025.emnlp-main.1542}.
\newblock URL \url{https://aclanthology.org/2025.emnlp-main.1542/}.

\bibitem[Pyatkin et~al.(2025)Pyatkin, Malik, Graf, Ivison, Huang, Dasigi,
  Lambert, and Hajishirzi]{pyatkin2025generalizing}
Valentina Pyatkin, Saumya Malik, Victoria Graf, Hamish Ivison, Shengyi Huang,
  Pradeep Dasigi, Nathan Lambert, and Hannaneh Hajishirzi.
\newblock Generalizing verifiable instruction following.
\newblock In \emph{Advances in Neural Information Processing Systems, Datasets
  and Benchmarks Track}, 2025.
\newblock URL \url{https://openreview.net/forum?id=yfYgwjj5F8}.

\bibitem[Qin et~al.(2024)Qin, Song, Hu, Yao, Cho, Wang, Wu, Liu, Liu, and
  Yu]{qin2024infobench}
Yiwei Qin, Kaiqiang Song, Yebowen Hu, Wenlin Yao, Sangwoo Cho, Xiaoyang Wang,
  Xuansheng Wu, Fei Liu, Pengfei Liu, and Dong Yu.
\newblock {InFoBench}: Evaluating instruction following ability in large
  language models.
\newblock In \emph{Findings of the Association for Computational Linguistics:
  ACL 2024}, pp.\  13025--13048. Association for Computational Linguistics,
  2024.
\newblock \doi{10.18653/v1/2024.findings-acl.772}.
\newblock URL \url{https://aclanthology.org/2024.findings-acl.772/}.

\bibitem[Qin et~al.(2025)Qin, Li, Li, Xu, Shi, Lin, Cui, Li, and
  Sun]{qin2025incentivizing}
Yulei Qin, Gang Li, Zongyi Li, Zihan Xu, Yuchen Shi, Zhekai Lin, Xiao Cui,
  Ke~Li, and Xing Sun.
\newblock Incentivizing reasoning for advanced instruction-following of large
  language models.
\newblock In \emph{Advances in Neural Information Processing Systems},
  volume~38, pp.\  108337--108401. Curran Associates, Inc., 2025.
\newblock URL
  \url{https://proceedings.neurips.cc/paper_files/paper/2025/hash/9baf31febefde7bd76023c2d2f13cbd7-Abstract-Conference.html}.

\bibitem[Schulman et~al.(2017)Schulman, Wolski, Dhariwal, Radford, and
  Klimov]{schulman2017proximal}
John Schulman, Filip Wolski, Prafulla Dhariwal, Alec Radford, and Oleg Klimov.
\newblock Proximal policy optimization algorithms, 2017.
\newblock URL \url{https://arxiv.org/abs/1707.06347}.

\bibitem[Shao et~al.(2024)Shao, Wang, Zhu, Xu, Song, Bi, Zhang, Zhang, Li, Wu,
  and Guo]{shao2024deepseekmath}
Zhihong Shao, Peiyi Wang, Qihao Zhu, Runxin Xu, Junxiao Song, Xiao Bi, Haowei
  Zhang, Mingchuan Zhang, Y.~K. Li, Y.~Wu, and Daya Guo.
\newblock {DeepSeekMath}: Pushing the limits of mathematical reasoning in open
  language models, 2024.
\newblock URL \url{https://arxiv.org/abs/2402.03300}.

\bibitem[Shenfeld et~al.(2026)Shenfeld, Pari, and
  Agrawal]{shenfeld2025rlsrazor}
Idan Shenfeld, Jyothish Pari, and Pulkit Agrawal.
\newblock {RL's Razor}: Why online reinforcement learning forgets less.
\newblock In \emph{The Fourteenth International Conference on Learning
  Representations}, 2026.
\newblock URL \url{https://openreview.net/forum?id=7HNRYT4V44}.

\bibitem[Song et~al.(2025)Song, Wei, Gao, Wang, Luo, Li, Yao, Xiong, Chen, Liu,
  and Wang]{song2025mitigating}
Feifan Song, Shaohang Wei, Bofei Gao, Yejie Wang, Wen Luo, Wei Li, Linli Yao,
  Weimin Xiong, Liang Chen, Tianyu Liu, and Houfeng Wang.
\newblock Mitigating overthinking through reasoning shaping, 2025.
\newblock URL \url{https://arxiv.org/abs/2510.09535}.

\bibitem[Song \& Zheng(2026)Song and Zheng]{song2026survey}
Mingyang Song and Mao Zheng.
\newblock A survey of on-policy distillation for large language models, 2026.
\newblock URL \url{https://arxiv.org/abs/2604.00626}.

\bibitem[Wang et~al.(2023)Wang, Wei, Schuurmans, Le, Chi, Narang, Chowdhery,
  and Zhou]{wang2022selfconsistency}
Xuezhi Wang, Jason Wei, Dale Schuurmans, Quoc Le, Ed~Chi, Sharan Narang,
  Aakanksha Chowdhery, and Denny Zhou.
\newblock Self-consistency improves chain of thought reasoning in language
  models.
\newblock In \emph{International Conference on Learning Representations}, 2023.
\newblock URL \url{https://openreview.net/forum?id=1PL1NIMMrw}.

\bibitem[Wei et~al.(2025)Wei, Li, Song, Luo, Zhuang, Tan, Guo, and
  Wang]{wei2025time}
Shaohang Wei, Wei Li, Feifan Song, Wen Luo, Tianyi Zhuang, Haochen Tan,
  Zhijiang Guo, and Houfeng Wang.
\newblock {TimE}: A multi-level benchmark for temporal reasoning of {LLM}s in
  real-world scenarios.
\newblock In \emph{Advances in Neural Information Processing Systems},
  volume~38, 2025.
\newblock URL
  \url{https://proceedings.neurips.cc/paper_files/paper/2025/hash/84f1e188c2be52f89f6e206bc37d092d-Abstract-Datasets_and_Benchmarks_Track.html}.
\newblock Datasets and Benchmarks Track.

\bibitem[Wen et~al.(2025)Wen, Zhong, Khan, Perez, Steinhardt, Huang, Bowman,
  He, and Feng]{wen2024mislead}
Jiaxin Wen, Ruiqi Zhong, Akbir Khan, Ethan Perez, Jacob Steinhardt, Minlie
  Huang, Samuel~R. Bowman, He~He, and Shi Feng.
\newblock Language models learn to mislead humans via {RLHF}.
\newblock In \emph{International Conference on Learning Representations}, 2025.
\newblock URL \url{https://openreview.net/forum?id=xJljiPE6dg}.

\bibitem[Wen et~al.(2026)Wen, Liu, Zheng, Ye, Wu, Wang, Xu, Liang, Li, Miao,
  Bian, and Yang]{wen2025rlvr}
Xumeng Wen, Zihan Liu, Shun Zheng, Shengyu Ye, Zhirong Wu, Yang Wang, Zhijian
  Xu, Xiao Liang, Junjie Li, Ziming Miao, Jiang Bian, and Mao Yang.
\newblock Reinforcement learning with verifiable rewards implicitly
  incentivizes correct reasoning in base {LLMs}.
\newblock In \emph{International Conference on Learning Representations}, 2026.
\newblock URL \url{https://openreview.net/forum?id=jGbRWwIidy}.

\bibitem[Wu et~al.(2026)Wu, Han, and Cai]{wu2026lightningopd}
Yecheng Wu, Song Han, and Hai Cai.
\newblock {Lightning OPD}: Efficient post-training for large reasoning models
  with offline on-policy distillation, 2026.
\newblock URL \url{https://arxiv.org/abs/2604.13010}.

\bibitem[Xing et~al.(2026)Xing, Wang, Gao, Li, and
  Tang]{xing2026trustregionopd}
Xingrun Xing, Haoqing Wang, Boyan Gao, Ziheng Li, and Yehui Tang.
\newblock Trust region on-policy distillation, 2026.
\newblock URL \url{https://arxiv.org/abs/2606.01249}.

\bibitem[Yang et~al.(2024)Yang, Zhang, Hui, Gao, Yu, Li, Liu, Tu, Zhou, Lin,
  Lu, Xue, Lin, Liu, Ren, and Zhang]{yang2024qwen25math}
An~Yang, Beichen Zhang, Binyuan Hui, Bofei Gao, Bowen Yu, Chengpeng Li,
  Dayiheng Liu, Jianhong Tu, Jingren Zhou, Junyang Lin, Keming Lu, Mingfeng
  Xue, Runji Lin, Tianyu Liu, Xingzhang Ren, and Zhenru Zhang.
\newblock {Qwen2.5-Math} technical report: Toward mathematical expert model via
  self-improvement, 2024.
\newblock URL \url{https://arxiv.org/abs/2409.12122}.

\bibitem[Yue et~al.(2025)Yue, Chen, Lu, Zhao, Wang, Yue, Song, and
  Huang]{yue2025does}
Yang Yue, Zhiqi Chen, Rui Lu, Andrew Zhao, Zhaokai Wang, Yang Yue, Shiji Song,
  and Gao Huang.
\newblock Does reinforcement learning really incentivize reasoning capacity in
  {LLMs} beyond the base model?
\newblock In \emph{Advances in Neural Information Processing Systems}, 2025.
\newblock URL \url{https://openreview.net/forum?id=4OsgYD7em5}.
\newblock Oral presentation.

\bibitem[Zhou et~al.(2023)Zhou, Lu, Mishra, Brahma, Basu, Luan, Zhou, and
  Hou]{zhou2023instruction}
Jeffrey Zhou, Tianjian Lu, Swaroop Mishra, Siddhartha Brahma, Sujoy Basu,
  Yi~Luan, Denny Zhou, and Le~Hou.
\newblock Instruction-following evaluation for large language models, 2023.
\newblock URL \url{https://arxiv.org/abs/2311.07911}.

\end{thebibliography}
\bibliographystyle{styles/iclr2027/report}
\appendix
\crefalias{section}{appendix}
\crefalias{subsection}{appendix}
\addtocontents{toc}{\protect\setcounter{tocdepth}{2}}
\clearpage
\phantomsection
\pdfbookmark[0]{Appendix}{appendix.contents}
\begingroup
\centering
{\ReportHeadingFont\fontsize{12}{14}\selectfont Appendix\par}
\endgroup
\vspace{12pt}
\begingroup
\hypersetup{linkcolor=reportCarnegieRed,linktoc=all}
\makeatletter
\renewcommand*{\l@section}[2]{%
  \addvspace{7pt}%
  {\bfseries\@dottedtocline{1}{0em}{1.8em}{#1}{\normalfont\hypersetup{linkcolor=black}#2}}}
\renewcommand*{\l@subsection}[2]{%
  \@dottedtocline{2}{1.8em}{2.4em}{#1}{\hypersetup{linkcolor=black}#2}}
\renewcommand*{\@dotsep}{2.5}
\setlength{\parskip}{3pt}
\edef\appendixsavedtocdepth{\number\value{tocdepth}}
\@starttoc{toc}
\setcounter{tocdepth}{\appendixsavedtocdepth}
\makeatother
\endgroup
\clearpage

\pretocmd{\section}{\FloatBarrier}{}{}
\section{Experimental Setup and Training Recipes}
\label{app:experimental_setup}
\label{app:experimental_details}

This appendix provides full training protocols, evaluation configurations, and analytical details for the experiments in the main text.

\subsection{Models and Base Policies}

We conduct experiments across two representative open-source model families: Qwen3-8B-Base and Qwen2.5-Math-7B.
For each model family, the base policy prior to any on-policy RLVR training is designated as Base and serves as the reference policy for distributional comparisons.
Training was conducted primarily on a node with eight NVIDIA H20 GPUs, each with 96 GB of GPU memory.

\subsection{Primary RLVR Training Paths}

Math-RLVR is trained on the public 7.5k training split of MATH, which spans a broad difficulty range and diverse problem types~\citep{hendrycks2021math}.
The math verifier uses an exact-match rule on extracted final boxed answers, granting positive rewards strictly to correct solutions.
IF-RLVR uses the IFTrain constraint library released with IFBench, whose 29 constraint types do not overlap with the 25 IFEval types~\citep{pyatkin2025generalizing,zhou2023instruction}.
The instruction verifier employs rule-based deterministic checkers to verify whether model outputs satisfy all explicit prompt constraints.
Each training and stochastic evaluation configuration is run once with the fixed random seed 42.
Repeated rollouts measure within-policy sampling variation rather than variation across independent training runs.

\subsection{Sequential Training and Mitigation Recipes}

The IF$\rightarrow$Math path initiates from the converged IF-RLVR endpoint and continues on-policy training with the math verifier.
The Math$\rightarrow$IF path initiates from the converged Math-RLVR endpoint and continues on-policy training with the instruction verifier.
In Math$\rightarrow$IF, the unconstrained run serves as a clean sequential baseline, while the reference-KL sweep uses \(\beta\in\{0,0.04,0.08,0.12\}\) to test the retention--adaptation trade-off.
\subsubsection{Routing-Prior Cold-Start Conditions}
\label{app:routing_prior_cold_start}

All four cold-start conditions first undergo SFT on 50 correct mathematical responses, while differing in opening composition and SFT configuration.
After SFT, all four policies receive the same 100-step IF-RLVR training with the vanilla b2r1 DAPO configuration.
The step-0 route statistics in the table are measured after SFT and before IF-RLVR.

\begin{table}[htbp]
  \centering
  \normalsize
  \setlength{\tabcolsep}{6pt}
  \renewcommand{\arraystretch}{1.15}
  \caption{
  \textbf{SFT configurations for the four routing-prior cold starts.}
  Soft DRI and Hard DRI share the same DRI-only corpus but use different SFT configurations, whereas the controls change the opening composition under the Hard DRI configuration.
  }
  \label{tab:cold_start_conditions}
  \begin{tabularx}{\linewidth}{@{}>{\raggedright\arraybackslash}p{2.3cm}*{3}{>{\raggedright\arraybackslash}X}@{}}
    \toprule
    \rowcolor{white}
    \textbf{Condition} & \textbf{SFT corpus} & \textbf{SFT configuration} & \textbf{Step-0 route statistics} \\
    \midrule
    Soft DRI (S1) &
    50 correct math responses; DRI-only openings (e.g., \texttt{To solve...}, \texttt{Let's...}) &
    Batch size 8; 10 epochs; learning rate \(2\times10^{-6}\); 60 gradient steps &
    \(\log p_{\mathrm{DRI}}=-0.17\); DAI rate \(=0\) \\
    \rowcolor{reportTableHighlight}
    Hard DRI (S2) &
    Same as Soft DRI &
    Batch size 32; 20 epochs; learning rate \(1\times10^{-5}\); 20 gradient steps &
    \(\log p_{\mathrm{DRI}}\approx 0\); DAI rate \(=0\) \\
    DAI control &
    50 correct math responses; DAI-only openings (e.g., \texttt{Answer:}, \texttt{The answer is...}) &
    Same as Hard DRI &
    DAI rate \(=1.000\); DRI rate \(=0\) \\
    \rowcolor{reportTableHighlight}
    Random control &
    50 correct math responses; mixed openings (\(\sim62\%\) DRI, \(30\%\) Other, \(8\%\) DAI) &
    Same as Hard DRI &
    DRI rate \(=0.885\); DAI rate \(=0\) \\
    \bottomrule
  \end{tabularx}
\end{table}

On-Policy Distillation aligns Base students to the on-policy rollout distributions of IF-RLVR teachers at steps 20, 40, 60, 80, or 100 using a top-$K$ reverse KL objective.

\section{Benchmark Suite and Decoding Protocols}
\label{app:benchmarks_and_protocols}

This section details the benchmark compositions, decoding configurations, and diagnostic metric formulations.

\subsection{Mathematics Benchmarks}

AIME24 and AIME25 each contain 30 high-difficulty competition mathematics problems, serving as high-difficulty validation sets for mathematical reasoning.
MATH-500-128 is a fixed 128-prompt subset stratified across difficulty levels 1 through 5 from MATH-500, probing mathematical support across a broader difficulty spectrum~\citep{hendrycks2021math}.

\subsection{Instruction-Following Benchmarks}

IFEval contains 541 prompts and 25 verifiable constraint types for evaluating basic precise instruction following~\citep{zhou2023instruction}.
IFBench applies 58 new out-of-domain constraints to 300 held-out WildChat prompts to test generalization to unseen constraints~\citep{pyatkin2025generalizing}.
The sources, relationships, and complete type inventories of the five constraint datasets are documented in \Cref{app:if_data_lineage}.
Official deterministic decoding settings report standardized benchmark scores, whereas repeated stochastic decoding (temperature 0.7, top-$p$ 0.95) diagnoses rewardable support within the policy distribution.

\subsection{Joint-Capability Benchmarks}

MathIF attaches one to three IFEval-style constraints to math problems and evaluates both answer correctness and full-response compliance~\citep{fu2025scaling}.
ReasonIF applies one constraint per prompt to the reasoning trace while placing the final answer in separate tags, thereby distinguishing reasoning-process following from final-response following~\citep{kwon2025reasonif}.

\subsection{Decoding Parameters and Support Metrics}

Core support diagnostics on AIME, IFEval, and IFBench use repeated stochastic decoding with 32 rollouts sampled per prompt.
Expected single-sample pass rate (\(\metricat{pass}{1}\)) estimates the expected success probability of the current policy under a single rollout.
Best-of-$k$ success rate (\(\metricat{best}{k}\)) measures whether at least one rewardable trajectory remains reachable within $k$ samples, reflecting the boundary of searchable support.
Pass-count records the number of strictly successful rollouts per prompt out of 32 (ranging from 0 to 32), partitioning prompts into all-wrong ($0/32$), mixed ($1\text{--}31/32$), and all-correct ($32/32$) support buckets.

\subsection{Support-Group Probes and Standalone Training Trajectories}
\label{app:standalone_support_trajectories}

For the analysis in \Cref{sec:starvation}, we sample repeated stochastic rollouts at each checkpoint over the same fixed set of math prompts.
A prompt is classified as all-wrong, mixed, or all-correct when none, some, or all of its sampled rollouts succeed, respectively.
These checkpoint probes are unfiltered and differ from the dynamic-sampling training logs, which do not produce policy gradients from prompts without within-group reward variation.
Because rollouts are resampled at each checkpoint, the plotted group shares are stochastic estimates rather than deterministic state counts.
The standalone IF-RLVR endpoint and the inherited IF$\rightarrow$Math step-0 state are probed separately, yielding all-wrong estimates of 64.1\% and 65.6\%, respectively.
We retain both measurements rather than forcing the sequential trajectory to start from the exact estimate shown for the standalone endpoint.

\begin{figure}[htbp]
  \centering
  \papergraphic{0.80\linewidth}{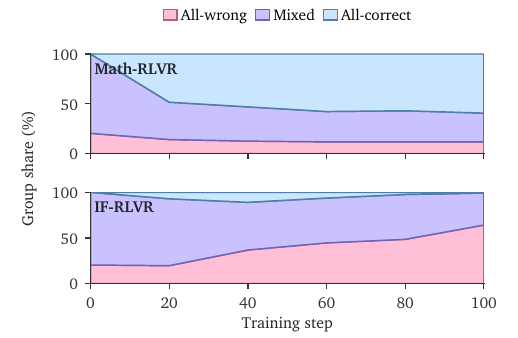}
  \caption{
  \textbf{Standalone Math-RLVR and IF-RLVR move math support-group composition in opposite directions.}
  \textbf{(a)} During Math-RLVR, the all-wrong share falls from 20.3\% to 11.7\% while the all-correct share expands.
  \textbf{(b)} During IF-RLVR, the all-wrong share rises from 20.3\% to 64.1\% while the all-correct share nearly vanishes.
  Filled areas connect independently probed checkpoints and do not represent measurements at intermediate training steps.
  }
  \label{fig:app_standalone_support_composition}
\end{figure}

These standalone trajectories show how the support-group compositions underlying the two reference paths in \Cref{fig:if2math_support_trajectory}a develop during training.
The sequential IF$\rightarrow$Math result remains a step-0-to-step-20 Sankey because that comparison uses only those two unfiltered checkpoint probes.

\section{Opening Route Classification and Behavioral Diagnostics}
\label{app:route_classifier}

This section provides exact pattern matching rules for the opening route classifier and the independent semantic judge.

\subsection{Deterministic Opening Route Classification Protocol}
\label{app:route_classifier_rules}

To ensure classification is strictly reproducible and independent of subjective interpretation, we construct a deterministic, non-parametric, priority-ordered prefix classifier.
The classifier operates exclusively on the initial 64-character window of each rollout, capturing opening-route decisions while preventing downstream reasoning steps from triggering opening-mode rules.

\begin{table}[htbp]
  \centering
  \normalsize
  \setlength{\tabcolsep}{6pt}
  \renewcommand{\arraystretch}{1.15}
  \begin{tabularx}{\linewidth}{@{}>{\centering\arraybackslash}p{1.5cm}>{\centering\arraybackslash}p{2.1cm}>{\raggedright\arraybackslash}p{2.7cm}X@{}}
    \toprule
    \rowcolor{white}
    \textbf{Priority} & \textbf{Route Mode} & \textbf{Target Behavior} & \textbf{Matching Prefix Patterns (Initial 64 Chars)} \\
    \midrule
    \textbf{P1} & \textbf{DAI} & Direct Answer Emission & \texttt{\textbackslash boxed\{\dots\}}, \texttt{"The answer is"}, \texttt{"Answer:"}, \texttt{"Final Answer:"}, bare numbers ($\le 10$ chars) \\
    \rowcolor{reportTableHighlight}
    \textbf{P2} & \textbf{CSI} & Constraint-Surface Shell & Front-loaded \texttt{"[placeholder]"} dumps, keyword headers, format shells (IF benchmarks) \\
    \textbf{P3} & \textbf{DRI} & Step-by-Step Deduction & \texttt{"To solve"}, \texttt{"Let's"}, \texttt{"Step 1:"}, \texttt{"We need to"}, \texttt{"Alright"}, \texttt{"First,"} \\
    \rowcolor{reportTableHighlight}
    \textbf{P4} & \textbf{Other} & Fallback Opening & Unmatched prefixes (guarantees mutually exclusive, exhaustive assignment) \\
    \bottomrule
  \end{tabularx}
  \caption{\textbf{Priority-cascaded protocol for deterministic opening route classification.} Rollouts are evaluated strictly in priority order (P1 $\rightarrow$ P4) over the initial 64 normalized characters.}
  \label{tab:route_classifier_protocol}
\end{table}

The classification pipeline executes text normalization prior to priority cascading.
Leading whitespace, newlines, and Markdown header symbols (e.g., \texttt{\#}, \textbf{**}) are stripped before matching.
Cascade evaluation stops at the first matching rule (P1 through P4), ensuring mutually exclusive and exhaustive route assignments.
Manual spot-checking across 500 randomly sampled rollouts confirms that this deterministic classifier achieves over 98.5\% precision, validating its reliability as an observable behavioral measure.

\subsection{Independent Semantic Judge (DeepSeek-V4-Pro)}
\label{apdx:shortcut_audit}

To audit the semantic quality of verifier-passed IF rollouts, an independent LLM judge evaluates sampled responses into a three-class semantic taxonomy.
Contentful responses provide thorough, genuine, and content-rich instruction-following outputs.
Shortcut responses pass deterministic verifiers by relying on superficial shortcuts such as brief text, templated prefixes, or placeholder dumps.
Others contains responses that fail verifier constraints or exhibit severe content defects.
Across training stages on a shared instruction-following prompt pool, shortcut responses expand from 4.5\% under the Base policy to 40.1\% at the final IF-RLVR checkpoint.
A focused audit of verifier-passed rollouts from the final checkpoint shows that 73.5\% rely on superficial shortcuts rather than substantive task fulfillment.

\section{Token-Level Mechanism and Distributional Analyses}
\label{app:token_mechanisms}

This section describes token-level divergence calculations and causal intervention methodologies.

\subsection{\texorpdfstring{Position-Wise JS Divergence over the Top-$K$ Union}{Position-Wise JS Divergence over the Top-K Union}}
\label{app:topk_divergence}

All distributional comparisons are made within one model lineage, so the Base, Math-RLVR, and IF-RLVR checkpoints share a vocabulary and tokenizer.
For a trace \((x_i,y_i)\) sampled from Base, all three policies are evaluated by teacher forcing on the same response prefix \(y_{i,<t}\).
\begin{equation}
p_{i,t}^{a}(v)=\pi_a\!\left(v\mid x_i,y_{i,<t}\right),\qquad a\in\{\mathrm B,\mathrm M,\mathrm I\}.
\label{eq:matched-prefix-distribution}
\end{equation}
Here \(t=1\) denotes the first generated token after the prompt and chat template.

For each comparison between Base and a checkpoint, we separately construct the two top-64 candidate sets and their union.
\begin{equation}
\begin{aligned}
S_{i,t}^{a}&=\operatorname{TopK}_{64}\!\left(p_{i,t}^{a}\right),\qquad a\in\{\mathrm B,c\},\\
U_{i,t}^{c}&=S_{i,t}^{\mathrm B}\cup S_{i,t}^{c},\qquad c\in\{\mathrm M,\mathrm I\}.
\end{aligned}
\label{eq:topk-union}
\end{equation}
For a token that appears only in the other policy's top-64 set, the implementation assigns zero mass under the current policy.
Each side is then renormalized over its retained top-64 probability mass.
\begin{equation}
\widetilde p_{i,t}^{a\mid c}(v)=
\begin{cases}
\dfrac{p_{i,t}^{a}(v)}{\sum_{u\in S_{i,t}^{a}}p_{i,t}^{a}(u)}, & v\in S_{i,t}^{a},\\[6pt]
0, & v\in U_{i,t}^{c}\setminus S_{i,t}^{a},
\end{cases}
\qquad a\in\{\mathrm B,c\}.
\label{eq:topk-renormalization}
\end{equation}
This construction is a zero-padded top-64-union approximation rather than a recovery of full-vocabulary probabilities for missing tokens.

To define the JS divergence reported in the main text, we first define KL divergence for normalized distributions \(P\) and \(Q\) over the same union \(U\).
\begin{equation}
\KL(P\Vert Q)=\sum_{v\in U}P(v)\log\frac{P(v)}{Q(v)}.
\label{eq:kl-definition}
\end{equation}
Let \(R=(P+Q)/2\); the Jensen--Shannon divergence is defined as follows.
\begin{equation}
D_{\mathrm{JS}}(P,Q)=\frac{1}{2}\KL(P\Vert R)+\frac{1}{2}\KL(Q\Vert R).
\label{eq:js-definition}
\end{equation}
The position-wise divergence in the main text is computed as \(\Delta_{i,t}^{c}=D_{\mathrm{JS}}(\widetilde p_{i,t}^{\mathrm B\mid c},\widetilde p_{i,t}^{c\mid c})\).
The tilde notation in the main text denotes this zero-padded and renormalized top-64-union comparison.
We use natural logarithms, so \(D_{\mathrm{JS}}\in[0,\log 2]\).

For \Cref{fig:js_opening_localization}, let \(n_i\) denote the number of tokens in response \(i\).
The six disjoint position groups are \(t=1\), \(2\leq t\leq4\), \(5\leq t\leq16\), \(17\leq t\leq63\), \(63<t\leq0.95n_i\), and \(t>\max(63,0.95n_i)\), respectively.
Interior and Tail 5\% denote the final two groups, respectively.
Each stem reports the token-level mean of \(\Delta_{i,t}^{c}\) over all tokens in its position group.
The figure uses the shared height mapping \(h=7\,\mathrm{pt}\sqrt{\overline{D}_{\mathrm{JS}}/\log 2}\).
The right-hand ratios instead compute position-group means within each trajectory, assign zero to a missing group, average each group across trajectories, and then take their ratio.
Thus, each trajectory has equal weight in the ratio, whereas the stem means are weighted by token count.

\begin{table}[!htbp]
  \centering
  \normalsize
  \setlength{\tabcolsep}{5pt}
  \renewcommand{\arraystretch}{1.10}
  \caption{
  \textbf{Absolute token-level mean JS values for the six position groups in \Cref{fig:js_opening_localization}.}
  Each row corresponds to one Base model, RLVR checkpoint, and benchmark combination.
  }
  \label{tab:js-position-values}
  \rowcolors{2}{reportTableStripe}{white}
  \begin{tabularx}{\linewidth}{@{}>{\raggedright\arraybackslash}p{0.20\linewidth}>{\raggedright\arraybackslash}p{0.17\linewidth}>{\raggedright\arraybackslash}p{0.12\linewidth}*{3}{>{\raggedleft\arraybackslash}X}@{}}
    \toprule
    \rowcolor{white}
    Base model & RLVR checkpoint & Benchmark & \(1\) & \(2\text{--}4\) & \(5\text{--}16\) \\
    \midrule
    Qwen3-8B-Base & Math-RLVR & AIME    & 0.595707 & 0.068857 & 0.049285 \\
      & & IFEval  & 0.139395 & 0.032068 & 0.021217 \\
      & & IFBench & 0.140402 & 0.032135 & 0.017383 \\
    \cmidrule(l){2-6}
      & IF-RLVR & AIME    & 0.595250 & 0.063240 & 0.035537 \\
      & & IFEval  & 0.431789 & 0.172884 & 0.111037 \\
      & & IFBench & 0.439830 & 0.182518 & 0.098056 \\
    \midrule
    Qwen2.5-Math-7B & Math-RLVR & AIME    & 0.133506 & 0.056596 & 0.045369 \\
      & & IFEval  & 0.174396 & 0.068909 & 0.034124 \\
      & & IFBench & 0.094763 & 0.021497 & 0.010524 \\
    \cmidrule(l){2-6}
      & IF-RLVR & AIME    & 0.679662 & 0.101265 & 0.034661 \\
      & & IFEval  & 0.650033 & 0.234947 & 0.109856 \\
      & & IFBench & 0.509209 & 0.132754 & 0.058436 \\
    \bottomrule
  \end{tabularx}

  \medskip

  \rowcolors{2}{reportTableStripe}{white}
  \begin{tabularx}{\linewidth}{@{}>{\raggedright\arraybackslash}p{0.20\linewidth}>{\raggedright\arraybackslash}p{0.17\linewidth}>{\raggedright\arraybackslash}p{0.12\linewidth}*{3}{>{\raggedleft\arraybackslash}X}@{}}
    \toprule
    \rowcolor{white}
    Base model & RLVR checkpoint & Benchmark & \(17\text{--}63\) & Interior & Tail 5\% \\
    \midrule
    Qwen3-8B-Base & Math-RLVR & AIME    & 0.029205 & 0.022043 & 0.036251 \\
      & & IFEval  & 0.014785 & 0.001533 & 0.001788 \\
      & & IFBench & 0.011781 & 0.001005 & 0.000650 \\
    \cmidrule(l){2-6}
      & IF-RLVR & AIME    & 0.014596 & 0.003586 & 0.009463 \\
      & & IFEval  & 0.081098 & 0.008762 & 0.010657 \\
      & & IFBench & 0.068664 & 0.004560 & 0.003156 \\
    \midrule
    Qwen2.5-Math-7B & Math-RLVR & AIME    & 0.020899 & 0.008970 & 0.022233 \\
      & & IFEval  & 0.022421 & 0.002797 & 0.012712 \\
      & & IFBench & 0.006918 & 0.001592 & 0.004721 \\
    \cmidrule(l){2-6}
      & IF-RLVR & AIME    & 0.011409 & 0.007879 & 0.025944 \\
      & & IFEval  & 0.071842 & 0.010818 & 0.037871 \\
      & & IFBench & 0.030475 & 0.005722 & 0.014427 \\
    \bottomrule
  \end{tabularx}
\end{table}

\begin{table}[htbp]
  \centering
  \normalsize
  \setlength{\tabcolsep}{6pt}
  \renewcommand{\arraystretch}{1.10}
  \caption{
  \textbf{Trajectory-pooled JS means underlying the four AIME ratios in \Cref{fig:js_opening_localization}.}
  Each trajectory has equal weight, and Ratio divides the position-1 mean by the Interior mean.
  }
  \label{tab:js-routing-localization}
  \rowcolors{2}{reportTableStripe}{white}
  \begin{tabularx}{0.82\linewidth}{@{}>{\raggedright\arraybackslash}p{0.20\linewidth}>{\raggedright\arraybackslash}p{0.19\linewidth}*{3}{>{\raggedleft\arraybackslash}X}@{}}
    \toprule
    \rowcolor{white}
    Base model & RLVR checkpoint & JS@1 & Interior JS & Ratio \\
    \midrule
    Qwen3-8B-Base & Math-RLVR & 0.596 & 0.0314 & \(19.0\times\) \\
      & IF-RLVR   & 0.595 & 0.0086 & \(69.6\times\) \\
    \cmidrule(l){2-5}
    Qwen2.5-Math-7B & Math-RLVR & 0.134 & 0.0136 & \(9.8\times\) \\
      & IF-RLVR   & 0.680 & 0.0064 & \(106.7\times\) \\
    \bottomrule
  \end{tabularx}
\end{table}

\begin{table}[htbp]
  \centering
  \normalsize
  \setlength{\tabcolsep}{3pt}
  \renewcommand{\arraystretch}{1.10}
  \caption{
  \textbf{Large token-level JS values are rare in the ten settings with complete coverage statistics.}
  Entries report the median token-level JS and the percentage of positions above each threshold.
  The available export does not contain coverage statistics for the two Qwen2.5-Math-7B comparisons on IFBench.
  }
  \label{tab:js-threshold-coverage}
  \rowcolors{2}{reportTableStripe}{white}
  \begin{tabularx}{\linewidth}{@{}>{\raggedright\arraybackslash}p{0.21\linewidth}>{\raggedright\arraybackslash}p{0.18\linewidth}>{\raggedright\arraybackslash}p{0.10\linewidth}*{3}{>{\raggedleft\arraybackslash}X}@{}}
    \toprule
    \rowcolor{white}
    Base model & RLVR checkpoint & Dataset & Median JS & JS \(>0.05\) (\%) & JS \(>0.1\) (\%) \\
    \midrule
    Qwen3-8B-Base & Math-RLVR & AIME    & \(3.68\times10^{-4}\) & 13.5 & 7.7 \\
      &                            & IFEval  & \(2.20\times10^{-5}\) & 0.9 & 0.3 \\
      &                            & IFBench & \(1.12\times10^{-5}\) & 0.4 & 0.1 \\
    \cmidrule(l){2-6}
      & IF-RLVR   & AIME    & \(4.51\times10^{-5}\) & 2.4 & 0.8 \\
      &                            & IFEval  & \(6.69\times10^{-5}\) & 6.7 & 3.4 \\
      &                            & IFBench & \(2.91\times10^{-5}\) & 3.2 & 1.5 \\
    \midrule
    Qwen2.5-Math-7B & Math-RLVR & AIME   & \(9.45\times10^{-5}\) & 5.2 & 2.1 \\
      &                            & IFEval & \(1.14\times10^{-4}\) & 1.8 & 0.8 \\
    \cmidrule(l){2-6}
      & IF-RLVR   & AIME   & \(3.46\times10^{-5}\) & 5.1 & 2.2 \\
      &                            & IFEval & \(3.45\times10^{-4}\) & 7.4 & 3.9 \\
    \bottomrule
  \end{tabularx}
\end{table}

Across the ten combinations of base model, RLVR checkpoint, and dataset with complete token-level coverage statistics, median JS ranges from \(1.1\times10^{-5}\) to \(3.7\times10^{-4}\).
Across combinations, the fraction of positions above \(0.05\) ranges from \(0.4\%\) to \(13.5\%\), and the fraction above \(0.1\) ranges from \(0.1\%\) to \(7.7\%\).

\subsection{Candidate-Set Overlap and Top-Token Promotion}

Each position stores the Base and RLVR top-64 candidates, and we compare the first 5, first 10, and all 64 candidates.
For the RLVR top-1 token \(R_1\), we record its rank within the Base top-64 and its Base probability after renormalization within that set.
When \(R_1\) is absent from the Base top-64, we record rank 65 and set this renormalized probability to zero.
The high-JS summary selects the top decile of position-wise JS within each model--RLVR--dataset setting, computes shared-candidate counts per position, and then averages them within the setting.
The position-1 summary first aggregates 32 rollouts within each AIME prompt and then summarizes 30 unique prompts in each row.
\Cref{tab:full-candidate-reuse} reports the complete high-JS summary for all twelve settings.

\begin{table}[!htbp]
\centering
\normalsize
\setlength{\tabcolsep}{4.5pt}
\renewcommand{\arraystretch}{1.10}
\caption{
\textbf{Complete candidate-reuse statistics at high-JS positions across all twelve settings.}
Each row averages over the top \(10\%\) of token positions by JS within one starting-policy, RLVR-training, and dataset setting.
The shared columns report the mean number of common tokens in the Base and RLVR top-5, top-10, and top-64 sets.
Base probability in the final column is renormalized within the Base top-64.
}
\label{tab:full-candidate-reuse}
\rowcolors{2}{reportTableStripe}{white}
\begin{tabularx}{\linewidth}{@{}>{\raggedright\arraybackslash}p{0.21\linewidth}>{\raggedright\arraybackslash}p{0.09\linewidth}>{\raggedright\arraybackslash}p{0.10\linewidth}*{3}{>{\raggedleft\arraybackslash}X}@{}}
\toprule
\rowcolor{white}
Starting policy & RLVR & Dataset & Shared top 5 & Shared top 10 & Shared top 64 \\
\midrule
Qwen3-8B-Base & Math & AIME & 3.50 & 6.88 & 43.28 \\
Qwen3-8B-Base & Math & IFEval & 4.44 & 8.92 & 57.44 \\
Qwen3-8B-Base & Math & IFBench & 4.49 & 8.94 & 57.34 \\
Qwen3-8B-Base & IF & AIME & 4.32 & 8.51 & 52.89 \\
Qwen3-8B-Base & IF & IFEval & 3.69 & 7.40 & 46.78 \\
Qwen3-8B-Base & IF & IFBench & 3.92 & 7.72 & 48.39 \\
\midrule
Qwen2.5-Math-7B & Math & AIME & 4.09 & 8.14 & 51.28 \\
Qwen2.5-Math-7B & Math & IFEval & 4.25 & 8.46 & 53.90 \\
Qwen2.5-Math-7B & Math & IFBench & 4.50 & 8.96 & 57.32 \\
Qwen2.5-Math-7B & IF & AIME & 4.00 & 7.80 & 49.72 \\
Qwen2.5-Math-7B & IF & IFEval & 3.61 & 7.06 & 44.27 \\
Qwen2.5-Math-7B & IF & IFBench & 4.06 & 8.00 & 51.24 \\
\bottomrule
\end{tabularx}

\medskip

\setlength{\tabcolsep}{3pt}
\rowcolors{2}{reportTableStripe}{white}
\begin{tabularx}{\linewidth}{@{}>{\raggedright\arraybackslash}p{0.20\linewidth}>{\raggedright\arraybackslash}p{0.08\linewidth}>{\raggedright\arraybackslash}p{0.09\linewidth}*{4}{>{\raggedleft\arraybackslash}X}@{}}
\toprule
\rowcolor{white}
Starting policy & RLVR & Dataset & \makecell{Base top-3\\reuse (\%)} & \makecell{Base top-64\\reuse (\%)} & \makecell{Median\\Base rank} & \makecell{Mean Base\\prob. (\%)} \\
\midrule
Qwen3-8B-Base & Math & AIME & 91.8 & 99.5 & 1 & 67.0 \\
Qwen3-8B-Base & Math & IFEval & 98.6 & 100.0 & 1 & 56.0 \\
Qwen3-8B-Base & Math & IFBench & 99.4 & 100.0 & 1 & 68.4 \\
Qwen3-8B-Base & IF & AIME & 98.3 & 100.0 & 1 & 70.9 \\
Qwen3-8B-Base & IF & IFEval & 90.0 & 99.2 & 1 & 52.9 \\
Qwen3-8B-Base & IF & IFBench & 96.0 & 99.6 & 1 & 66.5 \\
\midrule
Qwen2.5-Math-7B & Math & AIME & 96.3 & 99.8 & 1 & 58.8 \\
Qwen2.5-Math-7B & Math & IFEval & 97.3 & 99.8 & 1 & 58.4 \\
Qwen2.5-Math-7B & Math & IFBench & 99.3 & 100.0 & 1 & 59.2 \\
Qwen2.5-Math-7B & IF & AIME & 94.3 & 99.9 & 1 & 55.0 \\
Qwen2.5-Math-7B & IF & IFEval & 89.3 & 98.0 & 1 & 51.2 \\
Qwen2.5-Math-7B & IF & IFBench & 94.7 & 99.3 & 1 & 54.9 \\
\bottomrule
\end{tabularx}
\end{table}

\subsection{Single-Token and Prefix Interventions}
\label{app:opening_interventions}

Position-1 forced token intervention replaces the first generated token with the top opening token of the target policy before resuming free sampling.
Prefix forcing injects DRI or DAI opening prefixes to verify the causal control of response openings over downstream math searchability.

\section{Extended Results and Qualitative Cases}
\label{app:extended_results}

This section contains cross-family robustness validation and qualitative response comparisons.

\subsection{Cross-Model Robustness on Qwen2.5-Math-7B}

Experiments on Qwen2.5-Math-7B faithfully reproduce the core findings reported in the main text.
Math-RLVR consistently induces single-sample IF gains alongside best-of-32 contractions, while IF-RLVR causes a severe collapse in math searchability.

\subsection{Qualitative Case Studies}

Surface-compliant shortcut responses often front-load required placeholders or keywords in initial lines to satisfy rule checkers while omitting substantive content.
In contrast, substantive instruction-following rollouts from uncollapsed policies exhibit natural contextual flow and comprehensive task fulfillment.

\section{Joint-Capability Stress-Test Details}
\label{app:external_joint_validation}

\subsection{Endpoints and Sampling}
\label{app:external_joint_provenance}

This stress test uses an independent set of Qwen3-8B-Base, Math-RLVR step-720, and IF-RLVR step-720 endpoints rather than treating the latter two as the step-220/step-100 checkpoints used earlier.
For each benchmark, endpoint, and prompt, we sample 16 rollouts with temperature \(0.7\), top-\(p\) \(0.95\), and at most \(8{,}192\) new tokens.
Chapter 7 is therefore an independent joint-behavior stress test of the support-reshaping claim, not a continuous-checkpoint comparison with the earlier training trajectories.

\subsection{Metrics and Main-Table Filters}
\label{app:external_joint_metrics}

For each rollout, \(C\) denotes answer correctness, \(F\) denotes the benchmark-specific strict constraint event, and \(J=C\land F\) requires both events on the same rollout.
For MathIF, \(F\) requires the response to satisfy all supported constraints; for ReasonIF, \(F\) is the Reasoning IFS event on the reasoning trace.
The MathIF main table retains 290 prompts with scalar-form gold answers, complete constraint scores, and exactly 16 rollouts for every endpoint.
This fixed filter excludes answers that require symbolic equivalence under the current scoring package, prompts with incomplete constraint scores, and queries with conflicting rollout counts.
The ReasonIF main table excludes 10 language prompts unsupported by the constraint checker and uses source-aware correctness and Reasoning IFS on the remaining 290 prompts.
All results first compute binary events at the rollout level and then average within each benchmark; no pooled average is taken across benchmarks.
Changes reported as \% are computed from differences between unrounded endpoint means; we make no statistical-significance claim from these descriptive comparisons.

\section{Instruction-Constraint Data Lineage}
\label{app:if_data_lineage}

These datasets are not five independent taxonomies; they extend a common family of response constraints toward out-of-domain constraints, constrained mathematical answers, and constrained reasoning traces.
IFEval supplies the baseline verifiable types, IFTrain adds distinct training types, IFBench reserves another type set for out-of-domain testing, and MathIF and ReasonIF change the task content and constraint target~\citep{zhou2023instruction,pyatkin2025generalizing,fu2025scaling,kwon2025reasonif}.

\subsection{Roles, Sources, and Constraint Targets}

\begin{table}[htbp]
  \centering
  \normalsize
  \setlength{\tabcolsep}{4pt}
  \renewcommand{\arraystretch}{1.15}
  \caption{
  \textbf{Lineage of the instruction-constraint data used in this work.}
  A type denotes a verifier template rather than an instantiated keyword, count, or ending phrase; the Chapter 7 main-table filters are applied after these full benchmark sets are loaded.
  }
  \label{tab:if_data_lineage}
  \begin{tabularx}{\linewidth}{@{}>{\raggedright\arraybackslash}p{66pt}>{\raggedright\arraybackslash}p{118pt}>{\raggedright\arraybackslash}p{138pt}>{\raggedright\arraybackslash}X@{}}
    \toprule
    \rowcolor{white}
    \textbf{Data and role} & \textbf{Base prompt source} & \textbf{Constraint construction} & \textbf{Scored target} \\
    \midrule
    IF-RLVR train &
    Public IF-RLVR prompt construction over the Tulu 3.9 task mixture &
    Up to five instantiated constraints drawn from the 29 IFTrain verifier types used by our training path &
    Full response; every attached constraint must pass \\
    \rowcolor{reportTableHighlight}
    IFEval &
    541 author-constructed general prompts &
    25 types; 305/179/57 prompts contain one/two/three constraints &
    Full response under strict or loose deterministic checks \\
    IFBench &
    300 held-out WildChat prompts &
    58 OOD types; 256/44 prompts contain one/two constraints &
    Full response; the single-turn split is used here \\
    \rowcolor{reportTableHighlight}
    MathIF &
    AIME (60), GSM8K (90), MATH-500 (90), Minerva Math (90), and OlympiadBench (90) &
    15 IFEval-style types; 140 prompts each contain one/two/three constraints &
    Entire mathematical response plus answer correctness \\
    ReasonIF &
    AIME (61), AMC (54), ARC (59), GPQA (73), and GSM8K (53) &
    Six types; each of the 300 prompts contains one constraint &
    Reasoning trace; the final answer is isolated by answer tags \\
    \bottomrule
  \end{tabularx}
\end{table}

The separation of type inventories prevents treating IFBench as another split of IFEval.
The 25 IFEval types, 29 IFTrain types, and 58 IFBench types serve distinct roles as baseline evaluation, training coverage, and unseen-constraint evaluation.
All 15 MathIF types are IFEval-style response constraints, so MathIF changes the underlying task rather than the constraint paradigm.
ReasonIF instead applies familiar case, format, language, length, punctuation, and ending constraints to the reasoning trace rather than the final response.

\subsection{Complete Verifier-Type Inventory}

The following tables retain every public type identifier and state, in compact form, what its verifier checks.
Symbols in braces denote instance-level parameters whose values remain in each data file's \texttt{kwargs} or \texttt{constraint\_args} field.
Our IF-RLVR training path uses the 29 IFTrain types listed below; the public IF-RLVR recipe can additionally mix in the 25 IFEval types, while the 58 IFBench types remain reserved for unseen-constraint evaluation.

\begingroup
\ReportNeedspace{8\baselineskip}
\captionsetup{hypcap=false}
\captionof{table}{
\textbf{Constraints used by the IF-RLVR training set in this work (IFTrain; 29 types).}
The group and constraint columns together form the complete public type identifier.
}
\label{tab:constraints_iftrain}
\normalsize
\setlength{\LTleft}{0pt}
\setlength{\LTright}{0pt}
\setlength{\tabcolsep}{4pt}
\renewcommand{\arraystretch}{1.10}
\addtocounter{table}{-1}
\renewcommand{\theHtable}{constraints.iftrain}
\rowcolors{2}{reportTableStripe}{white}
\begin{longtable}{@{}>{\raggedright\arraybackslash}p{88pt}>{\raggedright\arraybackslash}p{123pt}>{\raggedright\arraybackslash}p{\dimexpr\textwidth-211pt-4\tabcolsep\relax}@{}}
\toprule
\rowcolor{white}
\textbf{Constraint group} & \textbf{Constraint} & \textbf{Description} \\
\midrule
\endfirsthead
\rowcolor{white}
\multicolumn{3}{@{}l}{\textit{IFTrain constraints (continued)}}\\
\toprule
\rowcolor{white}
\textbf{Constraint group} & \textbf{Constraint} & \textbf{Description} \\
\midrule
\endhead
\midrule
\rowcolor{white}
\multicolumn{3}{r@{}}{\textit{Continued on the next page}}\\
\endfoot
\bottomrule
\endlastfoot

\nolinkurl{copy} & \nolinkurl{repeat_phrase} & Repeat a supplied phrase \(N\) times, replacing one word on each repetition. \\
\addlinespace[1.5pt]
\nolinkurl{copy} & \nolinkurl{copy} & Copy the specified instruction verbatim instead of executing it. \\
\addlinespace[1.5pt]
\nolinkurl{new} & \nolinkurl{copy_span_idx} & Copy the text span delimited by character indices \(n_{\mathrm{start}}\) and \(n_{\mathrm{end}}\). \\
\addlinespace[1.5pt]
\nolinkurl{copy} & \nolinkurl{copying_simple} & Repeat the user request unchanged and do not answer it. \\
\addlinespace[1.5pt]
\nolinkurl{copy} & \nolinkurl{copying_multiple} & Repeat the request \(N\) times, separate copies with six asterisks, and do not answer it. \\
\midrule

\nolinkurl{first_word} & \nolinkurl{first_word_sent} & Start every sentence with the specified word. \\
\addlinespace[1.5pt]
\nolinkurl{first_word} & \nolinkurl{first_word_answer} & Start the response with the specified word. \\
\addlinespace[1.5pt]
\nolinkurl{last_word} & \nolinkurl{last_word_sent} & End every sentence, before punctuation, with the specified word. \\
\addlinespace[1.5pt]
\nolinkurl{last_word} & \nolinkurl{last_word_answer} & End the response with the specified word. \\
\midrule

\nolinkurl{keywords} & \nolinkurl{no_adjacent_consecutive} & Prevent adjacent words from starting with consecutive alphabet letters. \\
\addlinespace[1.5pt]
\nolinkurl{keywords} & \nolinkurl{word_once} & Include the specified keyword exactly once. \\
\addlinespace[1.5pt]
\nolinkurl{keywords} & \nolinkurl{word_count_different_numbers} & Make the specified word occur exactly \(N\) times. \\
\addlinespace[1.5pt]
\nolinkurl{keywords} & \nolinkurl{exclude_word_harder} & Exclude a specified keyword selected from the source instruction. \\
\addlinespace[1.5pt]
\nolinkurl{count} & \nolinkurl{lowercase_counting} & Allow each lowercase word to occur at most \(N\) times. \\
\addlinespace[1.5pt]
\nolinkurl{letters} & \nolinkurl{letter_counting} & Constrain the total number of letters relative to \(N\) using the requested relation. \\
\addlinespace[1.5pt]
\nolinkurl{letters} & \nolinkurl{letter_counting2} & Make the specified letter occur exactly \(N\) times. \\
\addlinespace[1.5pt]
\nolinkurl{count} & \nolinkurl{counting_composition} & Produce three divider-separated paragraphs with \(n_{\mathrm{sent}}\) sentences per paragraph and \(n_{\mathrm{words}}\) words per sentence. \\
\addlinespace[1.5pt]
\nolinkurl{count} & \nolinkurl{count_unique} & Use each word at most once in the response. \\
\addlinespace[1.5pt]
\nolinkurl{count} & \nolinkurl{count_increment_word} & Include the first keyword once and the second keyword twice. \\
\addlinespace[1.5pt]
\nolinkurl{keywords} & \nolinkurl{palindrome} & Include a palindrome in the response. \\
\addlinespace[1.5pt]
\nolinkurl{keywords} & \nolinkurl{keyword_specific_position} & Place the specified keyword at word \(m\) of sentence \(n\). \\
\addlinespace[1.5pt]
\nolinkurl{keywords} & \nolinkurl{start_end} & Start and end the response with the same word, with no trailing punctuation. \\
\midrule

\nolinkurl{detectable_format} & \nolinkurl{sentence_hyphens} & Join all sentences with hyphens and no intervening spaces. \\
\addlinespace[1.5pt]
\nolinkurl{detectable_format} & \nolinkurl{square_brackets} & Enclose every word in square brackets. \\
\addlinespace[1.5pt]
\nolinkurl{paragraphs} & \nolinkurl{paragraphs} & Produce two paragraphs separated by the specified Markdown divider. \\
\addlinespace[1.5pt]
\nolinkurl{paragraphs} & \nolinkurl{paragraphs2} & Produce exactly two paragraphs separated only by two line breaks. \\
\addlinespace[1.5pt]
\nolinkurl{detectable_format} & \nolinkurl{bigram_wrapping} & Wrap every consecutive word bigram in double angle quotation marks. \\
\addlinespace[1.5pt]
\nolinkurl{punctuation} & \nolinkurl{punctuation_dot} & Prohibit period characters throughout the response. \\
\addlinespace[1.5pt]
\nolinkurl{punctuation} & \nolinkurl{punctuation_exclamation} & Prohibit exclamation-mark characters throughout the response. \\

\end{longtable}
\endgroup

\begingroup
\ReportNeedspace{8\baselineskip}
\captionsetup{hypcap=false}
\captionof{table}{
\textbf{Complete IFEval constraint set (25 types).}
These constraints check the full response and form the type inventory of the baseline instruction-following evaluation.
}
\label{tab:constraints_ifeval}
\normalsize
\setlength{\LTleft}{0pt}
\setlength{\LTright}{0pt}
\setlength{\tabcolsep}{3.5pt}
\renewcommand{\arraystretch}{1.10}
\addtocounter{table}{-1}
\renewcommand{\theHtable}{constraints.ifeval}
\rowcolors{2}{reportTableStripe}{white}
\begin{longtable}{@{}>{\raggedright\arraybackslash}p{91pt}>{\raggedright\arraybackslash}p{108pt}>{\raggedright\arraybackslash}p{\dimexpr\textwidth-199pt-4\tabcolsep\relax}@{}}
\toprule
\rowcolor{white}
\textbf{Constraint group} & \textbf{Constraint} & \textbf{Description} \\
\midrule
\endfirsthead
\rowcolor{white}
\multicolumn{3}{@{}l}{\textit{IFEval constraints (continued)}}\\
\toprule
\rowcolor{white}
\textbf{Constraint group} & \textbf{Constraint} & \textbf{Description} \\
\midrule
\endhead
\midrule
\rowcolor{white}
\multicolumn{3}{r@{}}{\textit{Continued on the next page}}\\
\endfoot
\bottomrule
\endlastfoot

\nolinkurl{keywords} & \nolinkurl{existence} & Include every required keyword in the response. \\
\addlinespace[1.5pt]
\nolinkurl{keywords} & \nolinkurl{frequency} & Make the target word occur relative to \(N\) according to the specified count relation. \\
\addlinespace[1.5pt]
\nolinkurl{keywords} & \nolinkurl{forbidden_words} & Exclude every word in the supplied forbidden list. \\
\addlinespace[1.5pt]
\nolinkurl{keywords} & \nolinkurl{letter_frequency} & Make the target letter occur relative to \(N\) according to the specified count relation. \\
\midrule

\nolinkurl{language} & \nolinkurl{response_language} & Write the entire response only in the specified language. \\
\midrule

\nolinkurl{length_constraints} & \nolinkurl{number_paragraphs} & Produce \(N\) paragraphs separated by the required Markdown divider. \\
\addlinespace[1.5pt]
\nolinkurl{length_constraints} & \nolinkurl{number_words} & Constrain the response to at least, approximately, or at most \(N\) words. \\
\addlinespace[1.5pt]
\nolinkurl{length_constraints} & \nolinkurl{number_sentences} & Constrain the response to at least, approximately, or at most \(N\) sentences. \\
\addlinespace[1.5pt]
\nolinkurl{length_constraints} & \nolinkurl{nth_paragraph_first_word} & Produce \(N\) paragraphs and start paragraph \(i\) with the specified word. \\
\midrule

\nolinkurl{detectable_content} & \nolinkurl{postscript} & Append a postscript that starts with the specified marker. \\
\addlinespace[1.5pt]
\nolinkurl{detectable_content} & \nolinkurl{number_placeholders} & Include at least \(N\) square-bracket placeholders. \\
\midrule

\nolinkurl{detectable_format} & \nolinkurl{number_bullet_lists} & Produce exactly \(N\) Markdown bullet points. \\
\addlinespace[1.5pt]
\nolinkurl{detectable_format} & \nolinkurl{constrained_response} & Answer with exactly one item from the supplied option set. \\
\addlinespace[1.5pt]
\nolinkurl{detectable_format} & \nolinkurl{number_highlighted_sections} & Mark at least \(N\) sections with the required Markdown emphasis. \\
\addlinespace[1.5pt]
\nolinkurl{detectable_format} & \nolinkurl{multiple_sections} & Produce \(N\) sections and mark each section opening with the required splitter. \\
\addlinespace[1.5pt]
\nolinkurl{detectable_format} & \nolinkurl{json_format} & Format the entire response as JSON. \\
\addlinespace[1.5pt]
\nolinkurl{detectable_format} & \nolinkurl{title} & Include a title enclosed in double angle brackets. \\
\midrule

\nolinkurl{combination} & \nolinkurl{repeat_prompt} & Repeat the user request verbatim before answering it. \\
\addlinespace[1.5pt]
\nolinkurl{combination} & \nolinkurl{two_responses} & Give two different responses separated only by six asterisks. \\
\midrule

\nolinkurl{change_case} & \nolinkurl{english_capital} & Write the entire English response in uppercase letters. \\
\addlinespace[1.5pt]
\nolinkurl{change_case} & \nolinkurl{english_lowercase} & Write the entire English response in lowercase letters. \\
\addlinespace[1.5pt]
\nolinkurl{change_case} & \nolinkurl{capital_word_frequency} & Constrain the number of all-uppercase words relative to \(N\). \\
\midrule

\nolinkurl{startend} & \nolinkurl{end_checker} & End with the exact supplied phrase and place nothing after it. \\
\addlinespace[1.5pt]
\nolinkurl{startend} & \nolinkurl{quotation} & Enclose the entire response in double quotation marks. \\
\midrule

\nolinkurl{punctuation} & \nolinkurl{no_comma} & Prohibit commas throughout the response. \\

\end{longtable}
\endgroup

\begingroup
\ReportNeedspace{8\baselineskip}
\captionsetup{hypcap=false}
\captionof{table}{
\textbf{Complete IFBench out-of-distribution test constraints (58 types).}
Except for the custom group, these constraints are appended to held-out WildChat base prompts.
}
\label{tab:constraints_ifbench}
\normalsize
\setlength{\LTleft}{0pt}
\setlength{\LTright}{0pt}
\setlength{\tabcolsep}{3.5pt}
\renewcommand{\arraystretch}{1.10}
\addtocounter{table}{-1}
\renewcommand{\theHtable}{constraints.ifbench}
\rowcolors{2}{reportTableStripe}{white}
\begin{longtable}{@{}>{\raggedright\arraybackslash}p{53pt}>{\raggedright\arraybackslash}p{99pt}>{\raggedright\arraybackslash}p{\dimexpr\textwidth-152pt-4\tabcolsep\relax}@{}}
\toprule
\rowcolor{white}
\textbf{Constraint group} & \textbf{Constraint} & \textbf{Description} \\
\midrule
\endfirsthead
\rowcolor{white}
\multicolumn{3}{@{}l}{\textit{IFBench constraints (continued)}}\\
\toprule
\rowcolor{white}
\textbf{Constraint group} & \textbf{Constraint} & \textbf{Description} \\
\midrule
\endhead
\midrule
\rowcolor{white}
\multicolumn{3}{r@{}}{\textit{Continued on the next page}}\\
\endfoot
\bottomrule
\endlastfoot

\nolinkurl{count} & \nolinkurl{word_count_range} & Keep the response between the specified minimum and maximum word counts. \\
\addlinespace[1.5pt]
\nolinkurl{count} & \nolinkurl{unique_word_count} & Use at least \(N\) distinct words. \\
\addlinespace[1.5pt]
\nolinkurl{count} & \nolinkurl{conjunctions} & Use at least \(N\) different coordinating conjunctions. \\
\addlinespace[1.5pt]
\nolinkurl{count} & \nolinkurl{person_names} & Mention at least \(N\) distinct names from the verifier's fixed name list. \\
\addlinespace[1.5pt]
\nolinkurl{count} & \nolinkurl{numbers} & Include exactly \(N\) numerical items. \\
\addlinespace[1.5pt]
\nolinkurl{count} & \nolinkurl{punctuation} & Use every punctuation mark in the verifier's required inventory at least once. \\
\addlinespace[1.5pt]
\nolinkurl{count} & \nolinkurl{words_japanese} & Make every \(N\)-th word Japanese. \\
\addlinespace[1.5pt]
\nolinkurl{count} & \nolinkurl{pronouns} & Include at least \(N\) pronouns. \\
\addlinespace[1.5pt]
\nolinkurl{count} & \nolinkurl{keywords_multiple} & Make four specified keywords occur one, two, three, and five times, respectively. \\
\midrule

\nolinkurl{ratio} & \nolinkurl{stop_words} & Keep the stop-word share at or below the specified percentage. \\
\addlinespace[1.5pt]
\nolinkurl{ratio} & \nolinkurl{sentence_type} & Maintain a \(2{:}1\) count ratio of declarative to interrogative sentences. \\
\addlinespace[1.5pt]
\nolinkurl{ratio} & \nolinkurl{sentence_balance} & Use equal numbers of declarative, interrogative, and exclamatory sentences. \\
\addlinespace[1.5pt]
\nolinkurl{ratio} & \nolinkurl{overlap} & Keep the absolute difference between the trigram overlap ratio with the reference text and the target within \(2\%\). \\
\addlinespace[1.5pt]
\nolinkurl{ratio} & \nolinkurl{sentence_words} & Write three equal-character-length sentences without reusing words across them. \\
\midrule

\nolinkurl{words} & \nolinkurl{alphabet} & Cycle word initials through the alphabet, returning to A after Z. \\
\addlinespace[1.5pt]
\nolinkurl{words} & \nolinkurl{vowel} & Write one paragraph whose words use only one vowel type. \\
\addlinespace[1.5pt]
\nolinkurl{words} & \nolinkurl{consonants} & Give every word at least one cluster of consecutive consonants. \\
\addlinespace[1.5pt]
\nolinkurl{words} & \nolinkurl{palindrome} & Include at least ten palindromes of five or more characters each. \\
\addlinespace[1.5pt]
\nolinkurl{words} & \nolinkurl{prime_lengths} & Use only words whose character lengths are prime numbers. \\
\addlinespace[1.5pt]
\nolinkurl{words} & \nolinkurl{start_verb} & Start the response with a verb. \\
\addlinespace[1.5pt]
\nolinkurl{words} & \nolinkurl{repeats} & Repeat no word more than \(N\) times. \\
\addlinespace[1.5pt]
\nolinkurl{words} & \nolinkurl{odd_even_syllables} & Alternate words with odd and even numbers of syllables. \\
\addlinespace[1.5pt]
\nolinkurl{words} & \nolinkurl{last_first} & Reuse each sentence's final word as the next sentence's first word. \\
\addlinespace[1.5pt]
\nolinkurl{words} & \nolinkurl{paragraph_last_first} & Begin and end each paragraph with the same word. \\
\addlinespace[1.5pt]
\nolinkurl{words} & \nolinkurl{no_consecutive} & Prevent consecutive words from sharing an initial letter. \\
\addlinespace[1.5pt]
\nolinkurl{words} & \nolinkurl{keywords_specific_position} & Place the specified keyword at word \(m\) of sentence \(n\). \\
\addlinespace[1.5pt]
\nolinkurl{words} & \nolinkurl{words_position} & Use the specified keyword as both the second and penultimate word. \\
\midrule

\nolinkurl{sentence} & \nolinkurl{alliteration_increment} & Increase the number of alliterative words from each sentence to the next. \\
\addlinespace[1.5pt]
\nolinkurl{sentence} & \nolinkurl{keyword} & Include the specified keyword in sentence \(N\). \\
\addlinespace[1.5pt]
\nolinkurl{sentence} & \nolinkurl{increment} & Make each sentence exactly \(N\) words longer than the preceding sentence. \\
\midrule

\nolinkurl{format} & \nolinkurl{parentheses} & Nest parentheses, brackets, and braces to a depth of at least five. \\
\addlinespace[1.5pt]
\nolinkurl{format} & \nolinkurl{quotes} & Create at least three nested quotation levels while alternating quote styles. \\
\addlinespace[1.5pt]
\nolinkurl{format} & \nolinkurl{options} & Return one supplied option and no explanation. \\
\addlinespace[1.5pt]
\nolinkurl{format} & \nolinkurl{newline} & Put every word on a separate line. \\
\addlinespace[1.5pt]
\nolinkurl{format} & \nolinkurl{emoji} & End every sentence with an emoji. \\
\addlinespace[1.5pt]
\nolinkurl{format} & \nolinkurl{line_indent} & Increase indentation on each successive line to form a staircase. \\
\addlinespace[1.5pt]
\nolinkurl{format} & \nolinkurl{quote_unquote} & Follow every quoted span with an unquoted explanation. \\
\addlinespace[1.5pt]
\nolinkurl{format} & \nolinkurl{list} & Produce a non-bulleted list using the specified separator. \\
\addlinespace[1.5pt]
\nolinkurl{format} & \nolinkurl{thesis} & Start every section with a thesis statement marked as italic in HTML. \\
\addlinespace[1.5pt]
\nolinkurl{format} & \nolinkurl{sub-bullets} & Give every asterisk bullet at least one hyphen-marked sub-bullet. \\
\addlinespace[1.5pt]
\nolinkurl{format} & \nolinkurl{no_bullets_bullets} & Place at least two period-terminated sentences before at least two asterisk bullets. \\
\addlinespace[1.5pt]
\nolinkurl{format} & \nolinkurl{title_case} & Write the entire response in title case. \\
\addlinespace[1.5pt]
\nolinkurl{format} & \nolinkurl{output_template} & Fill the exact template containing Answer, Conclusion, and Future Outlook fields. \\
\addlinespace[1.5pt]
\nolinkurl{format} & \nolinkurl{no_whitespace} & Produce output with no whitespace characters. \\
\midrule

\nolinkurl{custom} & \nolinkurl{multiples} & Enumerate only multiples of seven while counting from 10 through 50. \\
\addlinespace[1.5pt]
\nolinkurl{custom} & \nolinkurl{mcq_count_length} & Generate four five-option art-history questions whose stems increase in length. \\
\addlinespace[1.5pt]
\nolinkurl{custom} & \nolinkurl{reverse_newline} & List African countries in reverse alphabetical order, one per line. \\
\addlinespace[1.5pt]
\nolinkurl{custom} & \nolinkurl{word_reverse} & Answer the fixed query with the response words in reverse order. \\
\addlinespace[1.5pt]
\nolinkurl{custom} & \nolinkurl{character_reverse} & Answer the fixed query with the response characters in reverse order. \\
\addlinespace[1.5pt]
\nolinkurl{custom} & \nolinkurl{sentence_alphabet} & Write 26 sentences whose first words progress from A through Z. \\
\addlinespace[1.5pt]
\nolinkurl{custom} & \nolinkurl{european_capitals_sort} & List European capitals above the stated latitude in descending latitude order. \\
\addlinespace[1.5pt]
\nolinkurl{custom} & \nolinkurl{csv_city} & Generate seven comma-delimited rows under the specified five-column city schema. \\
\addlinespace[1.5pt]
\nolinkurl{custom} & \nolinkurl{csv_special_character} & Generate fourteen CSV rows and quote one field containing a special character. \\
\addlinespace[1.5pt]
\nolinkurl{custom} & \nolinkurl{csv_quotes} & Generate three tab-delimited rows and enclose every field in double quotation marks. \\
\addlinespace[1.5pt]
\nolinkurl{custom} & \nolinkurl{date_format_list} & List the requested battle dates as comma-separated values in YYYY-MM-DD format. \\
\midrule

\nolinkurl{repeat} & \nolinkurl{repeat_change} & Repeat the request after changing its first word, and do not answer it. \\
\addlinespace[1.5pt]
\nolinkurl{repeat} & \nolinkurl{repeat_simple} & Ignore all other instructions and output only the supplied sentence. \\
\addlinespace[1.5pt]
\nolinkurl{repeat} & \nolinkurl{repeat_span} & Copy the text delimited by character indices \(n_{\mathrm{start}}\) and \(n_{\mathrm{end}}\). \\

\end{longtable}
\endgroup

\begingroup
\ReportNeedspace{8\baselineskip}
\captionsetup{hypcap=false}
\captionof{table}{
\textbf{Complete MathIF constraint set (15 types).}
These IFEval-style constraints check the complete mathematical response and are scored separately from answer correctness.
}
\label{tab:constraints_mathif}
\normalsize
\setlength{\LTleft}{0pt}
\setlength{\LTright}{0pt}
\setlength{\tabcolsep}{3.5pt}
\renewcommand{\arraystretch}{1.10}
\addtocounter{table}{-1}
\renewcommand{\theHtable}{constraints.mathif}
\rowcolors{2}{reportTableStripe}{white}
\begin{longtable}{@{}>{\raggedright\arraybackslash}p{91pt}>{\raggedright\arraybackslash}p{119pt}>{\raggedright\arraybackslash}p{\dimexpr\textwidth-210pt-4\tabcolsep\relax}@{}}
\toprule
\rowcolor{white}
\textbf{Constraint group} & \textbf{Constraint} & \textbf{Description} \\
\midrule
\endfirsthead
\rowcolor{white}
\multicolumn{3}{@{}l}{\textit{MathIF constraints (continued)}}\\
\toprule
\rowcolor{white}
\textbf{Constraint group} & \textbf{Constraint} & \textbf{Description} \\
\midrule
\endhead
\midrule
\rowcolor{white}
\multicolumn{3}{r@{}}{\textit{Continued on the next page}}\\
\endfoot
\bottomrule
\endlastfoot

\nolinkurl{keywords} & \nolinkurl{existence} & Include every required keyword in the mathematical response. \\
\addlinespace[1.5pt]
\nolinkurl{keywords} & \nolinkurl{frequency} & Make the target word occur relative to \(N\) according to the specified count relation. \\
\addlinespace[1.5pt]
\nolinkurl{keywords} & \nolinkurl{forbidden_words} & Exclude every word in the supplied forbidden list. \\
\midrule

\nolinkurl{language} & \nolinkurl{response_language} & Write the entire mathematical response only in the specified language. \\
\midrule

\nolinkurl{length_constraint_checkers} & \nolinkurl{number_words} & Constrain the response to at least, approximately, or at most \(N\) words. \\
\midrule

\nolinkurl{detectable_format} & \nolinkurl{number_bullet_lists} & Produce exactly \(N\) Markdown bullet points. \\
\addlinespace[1.5pt]
\nolinkurl{detectable_format} & \nolinkurl{number_highlighted_sections} & Mark at least \(N\) sections with the required Markdown emphasis. \\
\addlinespace[1.5pt]
\nolinkurl{detectable_format} & \nolinkurl{multiple_sections} & Produce \(N\) sections and mark each section opening with the required splitter. \\
\midrule

\nolinkurl{combination} & \nolinkurl{repeat_prompt} & Repeat the mathematical request verbatim before answering it. \\
\midrule

\nolinkurl{startend} & \nolinkurl{end_checker} & End with the exact supplied phrase and place nothing after it. \\
\addlinespace[1.5pt]
\nolinkurl{startend} & \nolinkurl{quotation} & Enclose the entire mathematical response in double quotation marks. \\
\midrule

\nolinkurl{change_case} & \nolinkurl{capital_word_frequency} & Constrain the number of all-uppercase words relative to \(N\). \\
\addlinespace[1.5pt]
\nolinkurl{change_case} & \nolinkurl{english_capital} & Write the entire English mathematical response in uppercase letters. \\
\addlinespace[1.5pt]
\nolinkurl{change_case} & \nolinkurl{english_lowercase} & Write the entire English mathematical response in lowercase letters. \\
\midrule

\nolinkurl{punctuation} & \nolinkurl{no_comma} & Prohibit commas throughout the mathematical response. \\

\end{longtable}
\endgroup

\begingroup
\ReportNeedspace{8\baselineskip}
\captionsetup{hypcap=false}
\captionof{table}{
\textbf{Complete ReasonIF constraint set (six types).}
Each verifier applies only to the reasoning trace; the final answer is placed in separate tags and is outside the constraint target.
}
\label{tab:constraints_reasonif}
\normalsize
\setlength{\LTleft}{0pt}
\setlength{\LTright}{0pt}
\setlength{\tabcolsep}{3.5pt}
\renewcommand{\arraystretch}{1.10}
\addtocounter{table}{-1}
\renewcommand{\theHtable}{constraints.reasonif}
\rowcolors{2}{reportTableStripe}{white}
\begin{longtable}{@{}>{\raggedright\arraybackslash}p{132pt}>{\raggedright\arraybackslash}p{98pt}>{\raggedright\arraybackslash}p{\dimexpr\textwidth-230pt-4\tabcolsep\relax}@{}}
\toprule
\rowcolor{white}
\textbf{Constraint group} & \textbf{Constraint} & \textbf{Description} \\
\midrule
\endfirsthead
\rowcolor{white}
\multicolumn{3}{@{}l}{\textit{ReasonIF constraints (continued)}}\\
\toprule
\rowcolor{white}
\textbf{Constraint group} & \textbf{Constraint} & \textbf{Description} \\
\midrule
\endhead
\midrule
\rowcolor{white}
\multicolumn{3}{r@{}}{\textit{Continued on the next page}}\\
\endfoot
\bottomrule
\endlastfoot

\nolinkurl{change_case} & \nolinkurl{english_capital} & Write the English reasoning trace in uppercase letters. \\
\addlinespace[1.5pt]
\nolinkurl{detectable_format} & \nolinkurl{json_format} & Format the reasoning trace as JSON. \\
\addlinespace[1.5pt]
\nolinkurl{language} & \nolinkurl{reasoning_language} & Write the entire reasoning trace only in the specified language. \\
\addlinespace[1.5pt]
\nolinkurl{length_constraint_checkers} & \nolinkurl{number_words} & Constrain the reasoning trace to at least or at most \(N\) words. \\
\addlinespace[1.5pt]
\nolinkurl{punctuation} & \nolinkurl{no_comma} & Prohibit commas throughout the reasoning trace. \\
\addlinespace[1.5pt]
\nolinkurl{startend} & \nolinkurl{end_checker} & End the reasoning trace with the exact supplied phrase and no later reasoning text. \\

\end{longtable}
\endgroup

\end{document}